\documentclass[10pt,twocolumn,letterpaper]{article}

\usepackage[pagenumbers]{cvpr} 

\usepackage{graphicx}
\usepackage{amsmath}
\usepackage{amssymb}
\usepackage{booktabs}
\usepackage{multirow}
\usepackage{makecell}
\usepackage{diagbox}
\usepackage[ title, titletoc]{appendix}

\usepackage[pagebackref,breaklinks,colorlinks]{hyperref}

\usepackage[capitalize]{cleveref}
\crefname{section}{Sec.}{Secs.}
\Crefname{section}{Section}{Sections}
\Crefname{table}{Table}{Tables}
\crefname{table}{Tab.}{Tabs.}

\def\confName{CVPR}
\def\confYear{2022}

\newcommand{\B}[1]{$\mathbf{#1}$}
\newcommand{\methodname}{CER-MVS}
\newcommand{\tnt} {Tanks-and-Temples} 
\newcommand{\dtu} {DTU}
\newcommand{\bld} {BlendedMVS}
\newcommand{\todo}[1]{}

\begin{document}

\title{Multiview Stereo with Cascaded Epipolar RAFT}

\author{Zeyu Ma\\
Princeton University\\
{\tt\small zeyum@princeton.edu}
\and
Zachary Teed\\
Princeton University\\
{\tt\small zteed@princeton.edu}
\and
Jia Deng\\
Princeton University\\
{\tt\small jiadeng@princeton.edu}
}

\maketitle


\begin{abstract}
We address multiview stereo (MVS), an important 3D vision task that reconstructs a 3D model such as a dense point cloud from multiple calibrated images. We propose \methodname{} (Cascaded Epipolar RAFT Multiview Stereo), a new approach based on the RAFT (Recurrent All-Pairs Field Transforms) architecture developed for optical flow. \methodname{} introduces five new changes to RAFT: epipolar cost volumes, cost volume cascading, multiview fusion of cost volumes, dynamic supervision, and multiresolution fusion of depth maps. \methodname{} is significantly different from prior work in multiview stereo. Unlike prior work, which operates by updating a 3D cost volume, \methodname{} operates by updating a disparity field. Furthermore, we propose an adaptive thresholding method to balance the completeness and accuracy of the reconstructed point clouds. Experiments show that our approach achieves competitive performance on \dtu{} (the second best among published results) and state-of-the-art performance on the Tanks-and-Temples benchmark (both the intermediate and advanced set). Code is available at \href{https://github.com/princeton-vl/CER-MVS}{https://github.com/princeton-vl/CER-MVS}

\end{abstract}

\section{Introduction}

Multiview stereo (MVS) is an important task in 3D computer vision. It seeks to reconstruct a full 3D model, typically in the form of a dense 3D point cloud, from multiple RGB images with known camera intrinsics and poses. It is a difficult task that remains unsolved; the main challenge is producing a 3D model that is not only accurate but also complete, that is, no parts should be missing and all fine details should be recovered. 

Many of the latest results of multiview stereo are achieved by deep networks. In particular, many recent leading methods~\cite{zhang2020visibility,yan2020dense} are variants of MVSNet~\cite{yao2018mvsnet}, a deep architecture that consists of two main steps: (1) constructing a 3D cost volume in the frustum of a reference view, by warping features from other views, and (2) using 3D convolutional layers to transform, or ``regularize'', the cost volume before using it to predict a depth map. The resulting depth maps, one from each reference view, are then combined to form a single 3D point cloud through a heuristic procedure. 

However, a drawback of MVSNet is that regularizing the 3D plane-sweeping cost volume using 3D convolutions can be costly in terms of computation and memory, potentially limiting the quality of reconstruction under finite resources. Subsequent variants~\cite{yao2019recurrent} of MVSNet have attempted to address this issue by replacing 3D convolutions with recurrent sequential processing of 2D slices.  Despite significant empirical improvements, however, such sequential processing can be suboptimal because the 3D cost volume does not have a natural sequential structure. 

In this work, we propose \methodname{}, a new deep-learning multiview stereo approach that is significantly different from existing methods. Like prior deep-learning work on multiview stereo, \methodname{} predicts individual depth maps and then fuses them, but differs significantly in how it predicts each depth map. Given a reference view and multiple neighbor views, \methodname{} constructs a 3D cost volume for each neighbor view by computing the similarity between each pixel in the reference view and pixels along the epipolar line, indexed by increments of inverse depth (i.e.\@ disparity) in the reference view. Then, the cost volumes from all neighbor views are aggregated into a single cost volume. \methodname{} uses a GRU to iteratively update a disparity field---the field that represents pixel correspondence. Each update is generated by the GRU by sampling from the aggregated cost volume using the current disparity field.

The key difference of \methodname{} from MVSNet and its variants lies in how depth is predicted from the 3D cost volume. MVSNet updates (i.e.\@ regularizes) the 3D cost volume and predicts depth through a soft argmax on the updated cost volume. In contrast, \methodname{} does not update the cost volume at all; instead it iteratively updates a disparity field, which is used to retrieve values from the cost volume. The final depth prediction is simply  the inverted disparity field. Updating a disparity field, which is less expensive than updating the cost volume,  can allow more effective use of finite computing resources. 

\methodname{} builds upon RAFT~\cite{teed2020raft}, an architecture that estimates optical flow between two video frames. Compared to RAFT, which cannot be directly applied to multiview stereo, \methodname{} introduces four novel changes: 
\begin{itemize}
\item \emph{Epipolar cost volume:} RAFT constructs a 4D cost volume that compares all pairs of pixels from two views, whereas we construct a 3D cost volume comparing each pixel in the reference view with pixels which are on the epipolar line in a neighbor view and spaced by uniform increments of disparity. 

\item \emph{Cost volume cascading:} Unlike RAFT, the size of our epipolar cost volumes depends not only on the image resolution but also the number of disparity increments. To reconstruct fine details, a large number of disparity increments is necessary, but can blow up GPU memory. To address this issue, we introduce cascaded epipolar cost volumes, a novel design in the context of RAFT. In particular, after a fixed number of RAFT iterations, we construct additional finer-grained epipolar cost volumes centered around current disparity predictions with finer increments of disparity, allowing reconstruction of fine details with less memory. 

\item \emph{Multiview fusion of cost volumes:} RAFT constructs a single cost volume from two views, whereas \methodname{} constructs multiple cost volumes, one for each neighbor of a reference view. The cost volumes are then aggregated into a single volume through a simple averaging operator. 

\item \emph{Dynamic supervision:} RAFT uses exponentially decaying weights to add up flow errors in each iteration. We also use such weights, but supervise a dynamic combination of depth errors and disparity errors.

\item \emph{Multiresolution fusion of depth maps:} RAFT operates on a single resolution of the input images, whereas \methodname{} applies the same network to predict depth maps on multiple resolutions, and aggregate the depth maps into a single high-resolution depth map through a simple but novel heuristic. 
\end{itemize}

When stitching the depth maps into point clouds, a filtering algorithm is often used, e.g., Dynamic Consistency Checking proposed in D2HC-RMVSNet~\cite{yan2020dense}. However, a good balance of accuracy and completeness is required for high scores on the evaluation metric, which is ignored by these algorithms. Therefore, we propose an adaptive thresholding method built on top of ~\cite{yan2020dense}.

We evaluate \methodname{} on two challenging benchmarks, \dtu{}~\cite{aanaes2016large} and \tnt{}~\cite{knapitsch2017tanks}. On \dtu{}, \methodname{} achieves performance competitive to the current state of the art (the second best among published results). On \tnt{}, \methodname{} significantly advances the state of the art of the intermediate set from a mean F1 score of $61.68$ to $64.82$, and the advanced set from $37.44$ to $40.19$.

\begin{figure*}[t]
\begin{center}
    \includegraphics[width=\textwidth]{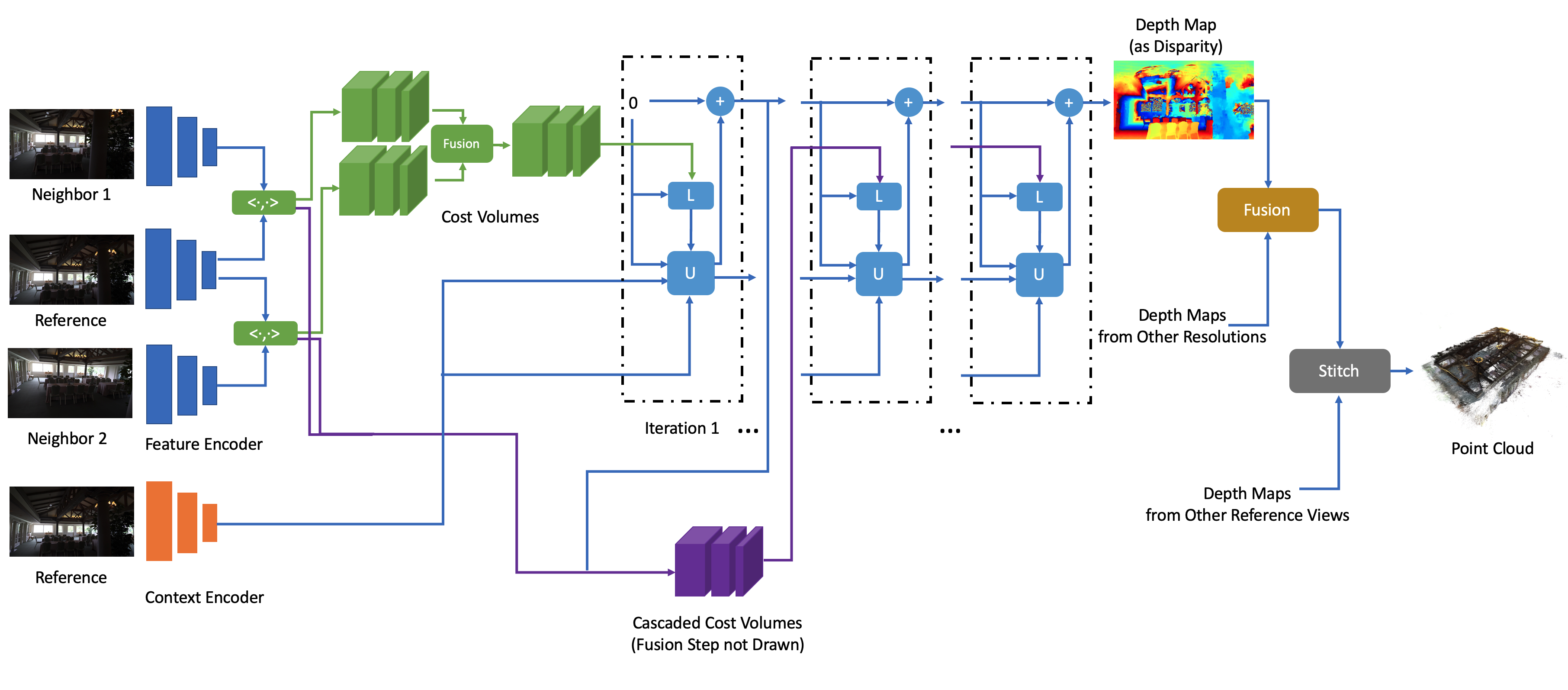}
\end{center}
  \caption{Overview of \methodname{}, which includes an architecture that constructs cascaded epipolar cost volumes and performs recurrent iterative updates of disparity (inverse depth) maps, with fusion of cost volumes from multiple views as well as fusion of disparity maps of multiple resolutions. }
  \label{fig:pipeline}
\end{figure*}

\section{Related Work}

\paragraph{Classical MVS}
Classical methods~\cite{campbell2008using,furukawa2009accurate,galliani2015massively,schonberger2016pixelwise,tola2012efficient,hirschmuller2007stereo} essentially formulate multiview stereo as an optimization problem, which seeks to find a 3D model that is most compatible with the observed images. The compatibility is typically based on some hand-designed notion of photo-consistency, assuming that pixels that are projections of the same 3D point should have similar appearance. Often photo-consistency alone does not sufficiently constrain the solution space, and the optimization objective can also include shape priors, which make additional assumptions about what shapes are likely. To solve the optimization problem, a concrete classical algorithm usually consists of a particular 3D representation (e.g.\@ polygon meshes, voxels, or depth maps) and a optimization procedure to compute the best model under that representation. The different combinations of photo-consistency measures, shape priors, 3D representations, and optimization procedures give rise to a large variety of algorithms. For more details, we refer the reader to excellent surveys of these algorithms by Seitz et al.~\cite{seitz2006comparison} and  by Furukawa and Hern\'{a}ndez ~\cite{10.1561/0600000052}.

One family of classical MVS methods~\cite{zheng2014patchmatch,schonberger2016pixelwise,galliani2015massively,xu2019multi,xu2020planar,romanoni2019tapa} is based on the PatchMatch~\cite{barnes2009patchmatch} algorithm, which enables efficient dense matching of pixels across views. PatchMatch methods have proved very effective and have demonstrated highly competitive performance. In particular, Xu and Tao~\cite{xu2020planar} introduced the ACMP algorithm, which, among other enhancements, incorporates planar priors and has achieved competitive results on \tnt{}. 

\paragraph{Learning-based MVS}
Unlike classical algorithms, our approach is learning-based. Existing learning-based MVS methods either use learning to improve parts of a classical pipeline such as PatchMatch \cite{zagoruyko2015learning,han2015matchnet,zbontar2015computing,zbontar2016stereo}, or develop end-to-end architectures~\cite{kar2017learning,ji2017surfacenet,yao2018mvsnet,yao2019recurrent,chen2019point,luo2019p,xue2019mvscrf,gu2020cascade,yang2020cost,yu2020fast,xu2020learning,yi2020pyramid,cheng2020deep,yan2020dense,zhang2020visibility}. A common step in existing end-to-end architectures is the construction of a 3D cost volume (or feature grid) through some differentiable geometric operations.  Then, this 3D cost volume undergoes further updates, often through 3D convolutions, before being transformed into the final 3D model in some particular representation such as voxels~\cite{kar2017learning,ji2017surfacenet}, depth maps~\cite{yao2018mvsnet,yao2019recurrent,luo2019p,xue2019mvscrf,gu2020cascade,yang2020cost,yu2020fast,xu2020learning,yi2020pyramid,cheng2020deep,yan2020dense,zhang2020visibility,ma2021epp,wei2021aa}, or point clouds~\cite{chen2019point}.

The main difference between our approach and existing works is that although we also construct a 3D cost volume, we do not update it. Instead, we update an inverse-depth field that is used to iteratively index from the 3D cost volume to produce 2D feature maps. Our approach thus avoids the costly operations of updating a 3D volume and focuses limited computing resources on refining the depth maps directly.

\section{Approach}

This section describes the detailed architecture and pipeline of \methodname{}, as shown in Fig.~\ref{fig:pipeline}. Given a reference view and a set of neighbor views, we first extract features using a set of convolutional networks. Features are then used to build a collection of cost volumes. We then predict a depth map through recurrent iterative updates, followed by the fusion of multiresolution depths. Finally, depth maps from all references views are fused and stitched to produce a final point cloud.

\subsection{Cost Volume Construction}

\newcommand{\Hf} {H_{\rm f}} 
\newcommand{\Wf} {W_{\rm f}}
\newcommand{\Df} {D_{\rm f}} 
\newcommand{\f} {\mathbf{f}} 
\newcommand{\N} {N} 
\newcommand{\C} {\mathbf{C}} 
\newcommand{\DD} {D} 
\newcommand{\CP} {\mathbf{C}_{\rm P}} 
\newcommand{\LL} {L} 
\newcommand{\fine} {^{\rm f} } 
\newcommand{\Kone} {T_1} 
\newcommand{\Ktwo} {T_2} 
\newcommand{\disp} {\mathbf{d}} 
\newcommand{\h} {\mathbf{h}} 
\newcommand{\Dh} {D_{\rm h}} 
\newcommand{\inp} {\mathbf{i}} 
\newcommand{\corr} {\mathbf{c}} 
\newcommand{\x} {\mathbf{x}} 
\newcommand{\z} {\mathbf{z}} 
\newcommand{\rr} {\mathbf{r}} 

\paragraph{Image Features}

We need to extract image features from both reference views and neighbor views before using them to construct the cost volumes. In addition,  the iterative update unit, to be introduced later, needs context features from reference views. We extract these image features using convolutional encoders following RAFT: $\mathbb{R} ^ {H \times W \times 3} \rightarrow \mathbb{R} ^ {H / 2^k \times W / 2^k \times \Df} $, where $k$ and $\Df$ are hyperparameters that control the feature resolution and dimension (See Sec.~\ref{sec:imp} and Appendix ~\ref{appA}  for more details).

\paragraph{Epipolar Cost Volume}

After extracting feature maps $\{\f_i, i=0,...,N+1\}$, where $\f_0$ is the reference view and others are neighbor views, each with resolution $(\Df, \Hf, \Wf) = (\Df, H / 2^k, W / 2^k)$, we construct a 3D cost volume by computing the correlation of each pixel in the reference view with pixels along its epipolar line in a neighbor view.  Specifically, for a pixel in the reference view, we backproject it to $\DD$ 3D points with disparity (inverse depth) uniformly spaced in the range from $0$ to $d_{\rm max}$ (after proper scaling as described in Sec.~\ref{sec:imp}), reproject the 3D points to the epipolar line in the neighbor view, and use differentiable bilinear sampling to retrieve the features from the neighbor view. This procedure outputs a volume $\C \in  \mathbb{R} ^ {\N \times \Hf \times \Wf \times \DD}$.

Like RAFT, we compute a stack of $\CP$ of multiscale cost volumes by repeated average-pooling, i.e., $\CP = \{\C _0, \C _1, ..., \C _{\LL -1}\}$ where $\C _l \in  \mathbb{R} ^ {N \times \Hf \times \Wf \times \DD / 2^l}, \text{ for } l =0, ..., \LL -1$.

\paragraph{Cost Volume Cascading}
Unlike RAFT, the size of an epipolar cost volume depends on not only the image resolution but also the number of disparity values sampled. A dense sampling of a large number of disparity values effectively increases the resolution of the cost volume along the depth dimension and can help reconstruct fine details. However, using a large number of disparity values can take too much GPU memory. To address this issue, we introduce a cascade design. The basic idea is to construct additional cost volumes that are finer-grained along the disparity dimension and centered around the current disparity predictions.

Concretely, after $\Kone$ iterative updates, we create a new stack of cost volumes $\CP \fine = \{\C \fine _0, \C \fine _1, ..., \C \fine _{\LL -1}\}$, $\C \fine _l \in  \mathbb{R} ^ {\N \times \Hf \times \Wf \times \DD \fine  / 2^l}, l =0, ..., \LL -1$, where  $ \DD \fine $ is the number of disparity values uniformly sampled centered around the current prediction of disparities with smaller increments than those used in the initial stack of cost volumes. Specifically, the value of $\DD \fine$ is determined by $2^{L - 1} * R$, where $R$ is a hyperparameter that controls the size of the neighborhood described in Sec. \ref{sec:iter}. The factor $2^{L-1}$ is needed to allow repeated pooling. In this work we use up to 2 stages in our experiments, but the design can be trivially extended to more stages. 

It is worth noting that cost volume cascading has been used in prior MVS work~\cite{gu2020cascade,yang2020cost}, but it is a novel design in the context of a RAFT-like architecture, which differs significantly from prior MVS work in that the cost volumes are not updated and are only used as static lookup tables.

\subsection{Iterative Updates}
\label{sec:iter}
The iterative updates follow RAFT in overall structure.  We iteratively update a disparity field $\disp \in \mathbb{R} ^ {\Hf \times \Wf}$  initialized to zero. In each iteration, the input to the update operator includes a hidden state $\h \in \mathbb{R} ^ {\Hf \times \Wf \times \Dh }$, the current disparity field,   the context features $\inp \in \mathbb{R} ^ {\Hf \times \Wf \times \Dh }$ from the reference view, as well as per-pixel features retrieved from the cost volumes using the current disparity field. The output of the update operator includes a new hidden state and an increment to the disparity field.

\paragraph{Multiview Fusion of Cost Volumes} Different from RAFT, in multiview stereo we need to consider multiple neighbor views. For each pixel in the reference view, we generate one correlation feature vector against each neighbor view. Given such feature vectors from multiple neighbor views, we take the element-wise mean as the final vector. The intuition behind this operator is that mean value is more robust as the number of neighbor views can vary in test time. 

To generate the correlation feature vector for each pixel against a single neighbor view, we perform the same lookup procedure as RAFT. Given the current disparity estimate for the pixel and the stack of cost volumes $\CP=\{\C _0, \C _1, ..., \C _{\LL -1}\}$ against the neighbor views, we retrieve, from each cost volume, correlation values corresponding to
 a local 1D integer grid of length $R$ centered around the current disparity. This is repeated for each level of the stack, and the values from all levels are concatenated to form a single feature vector.

\paragraph{Update Operator} We use a GRU-based update operator to propose a sequence of incremental updates to the disparity field.

First, we extract features from the current disparity estimate $\disp _t$. The feature vector is formed by subtracting the disparity of each pixel by its 7x7 neighborhood, then reshaping the result into a 49-dimensional vector. This operation has the effect of making the feature vector invariant to the disparity field up to a shift factor, since the retrieved vector only depends on relative disparity between neighboring pixels.

Second, because we have a cascade of cost volumes and our update operator accesses different cost volumes at different stages of the cascade, the operator, while still recurrent, should be given the flexibility to behave somewhat differently for different stages of the cascade. Thus, we modify the weight tying scheme of RAFT such that some weights are tied across all iterations while others are tied only within a single stage of the cascade. Specially, we tie all weights across iterations except the decoder layer that decodes a disparity update from the hidden state of the GRU. The weights of the decoder layer are tied only within each stage of the cascade.

Third, RAFT uses upsampling layers for final predictions of flow field, whereas we do not use any upsampling layer.

The update equations are as follows, with a 2-stage cascade with $T_1$ iterations for stage 1.  
\begin{align}
\x_{t} &= [ {\rm Encoder}_d(\disp _t), {\rm Encoder}_c(\corr), \inp] \\
\z_{t} &=\sigma\left(\operatorname{Conv}_{3 \times 3}\left(\left[\h_{t-1}, \x _{t} \right], W_{z}\right)\right) \\
\rr_{t} &=\sigma\left(\operatorname{Conv}_{3 \times 3}\left(\left[\h_{t-1}, \x _{t} \right], W_{r}\right)\right) \\
\tilde{\h}_{t} &=\tanh \left(\operatorname{Conv}_{3 \times 3}\left(\left[\rr _{t} \odot \h_{t-1}, \x _{t}\right], W_{h}\right)\right) \\
\h_{t} &=\left(1-\z _{t}\right) \odot \h_{t-1}+\z _{t} \odot \tilde{\h}_{t}\\
\Delta \disp _t &= \left\{
             \begin{array}{lr}
             {\rm Decoder}_1 (\h_t), & t \le \Kone \\
             {\rm Decoder}_2 (\h_t), & t > \Kone
             \end{array}
\right.
\end{align}
Here $\inp$ is the context features, and ${\rm Encoder}_c$ is an encoder the transforms the correlation features using two convolution layers (see  Appendix ~\ref{appA}  for details).

\subsection{Multiresolution Depth Fusion}
To construct fine details, it generally helps to operate at high resolution, but the available GPU memory limits the highest resolution the network can access, especially during training with large mini-batches. One approach to get around this limit is to apply the network to a higher resolution during inference, which is the common approach adopted in prior works. 

However, we find that while using a higher resolution during inference can help, an even better approach is to apply the same network on two input resolutions, the ``low'' resolution $W\times H$ used to train the network and the higher resolution $2W \times 2H$, and combine the two disparity maps ${\rm LR}$ and $\rm HR$ to form a fused disparity map ${\rm MR}$ with a control parameter $t$: 

\begin{equation}
\mathbf{d}_{\rm MR}=
\left\{
             \begin{array}{lr}
             \disp _{\rm HR}, &  {\rm if} |\disp_{\rm LR}^{-1} - \disp_{\rm HR}^{-1}| < t * \disp _{\rm LR}^{-1} \\
             \disp _{\rm LR}, &  {\rm otherwise}
             \end{array}
\right.
\end{equation}
That is, if the low resolution prediction and high resolution prediction are similar at a pixel, we use the high resolution prediction; otherwise we use the low resolution prediction. 
This is motivated by the observation that low resolution predictions are more reliable in term of texture-less large structures such as planes, whereas high resolution predictions are more reliable in terms of fine details, which do not tend to deviate drastically from low resolution predictions. 
Note that as the control parameter $t$ varies from 0 to infinity, $\disp_{\rm MR}$ varies from $\disp_{\rm LR}$ to $\disp _{\rm HR}$.

\begin{table}[]
\caption{Implementation hyperparameters}
\label{tab:imp}

\resizebox{\linewidth}{!}{

\begin{tabular}{l|cc}
\hline
Training dataset                & \multicolumn{1}{c|}{DTU}                                  & BlendedMVS                           \\ 
\hline
Native resolution  $(H, W)$             & \multicolumn{1}{c|}{(1200, 1600)}                         & (1536, 2048)                         \\ \hline
\# neighbor views               & \multicolumn{1}{c|}{10}                                   & 8                                    \\ \hline
\# training epochs              & \multicolumn{1}{c|}{15}                                   & 16                                   \\ \hline

Feature map downsize ratio      & \multicolumn{2}{c}{4}                                                                            \\ \hline
Feature map dimension           & \multicolumn{2}{c}{64}                                                                           \\ \hline
Cost volume stack size  $L$         & \multicolumn{2}{c}{3}                                                                           \\ \hline
Retrieved neighborhood size  $R$         & \multicolumn{2}{c}{11}                                                                           \\ \hline
Cascaded stages                 & \multicolumn{2}{c}{2}                                                                            \\ \hline
Max disparity $d_{\rm max}$    & \multicolumn{2}{c}{0.0025}                                                                            \\ \hline
Disparity increment in stage 1  & \multicolumn{2}{c}{$d_{\rm max}$ / 64}                                                                  \\ \hline
Disparity increment in stage 2  & \multicolumn{2}{c}{$d_{\rm max}$ / 320}                                                                 \\ \hline
\# GRU iterations in each stage & \multicolumn{2}{c}{8}                                                                            \\ \hline
Batch size                      & \multicolumn{2}{c}{2}                                                                            \\ \hline
Loss parameter                  & \multicolumn{2}{c}{$\lambda = 2.8\times 10 ^ {-6} , \kappa = 100, \gamma=0.9$} \\ \hline

\end{tabular}
}

\resizebox{\linewidth}{!}{
\begin{tabular}{l|c|c}
\hline
Test dataset                                                                                  & DTU          & Tanks-and-Temples                                                      \\ \hline
Native resolution       $(H, W)$                                                                       & (1200, 1600) & \begin{tabular}[c]{@{}c@{}}(1080, 1920)\\ or (1080, 2048)\end{tabular} \\ \hline
\begin{tabular}[c]{@{}l@{}}\# Neighbor views\\ for native resolution   input\end{tabular}     & 10          & 15                                                                    \\ \hline
\begin{tabular}[c]{@{}l@{}}\# Neighbor views\\ for 2 $\times$ native   resolution input\end{tabular} & 10          & 25                                                                    \\ \hline
\begin{tabular}[c]{@{}l@{}}Multires fusion\\ threshold $t$ \end{tabular}         & 0.02          & 0.02                                                                    \\ \hline
\begin{tabular}[c]{@{}l@{}}Resolution\\ for point cloud stitching\end{tabular}                &  native resolution          &           1/2    native resolution                                                       \\ \hline
\begin{tabular}[c]{@{}l@{}}Adaptive thresholding\\ parameter $p$ \end{tabular}                   & 0.25          & 0.25                                                                    \\ \hline
\end{tabular}
}

\end{table}

 \begin{table}

    \caption{Results on \textit{DTU} test set}
    
    \centering
    \label{tab:dtu}

  \begin{tabular}{lccc}
\hline
     & \multicolumn{3}{c}{DTU   mean distance (mm)} \\ \cline{2-4} 
     & Acc.         & Comp.        & Overall        \\ \hline
            COLMAP \cite{schonberger2016pixelwise}                &$0.400$       &$0.664$       &$0.532$   \\
            MVSNet \cite{yao2018mvsnet}                           &$0.396$       &$0.527$       &$0.462$   \\
            D2HC-MVSNet \cite{yan2020dense}                       &$0.395$       &$0.378$       &$0.386$   \\ 
            Point-MVSNet \cite{chen2019point}                     &$0.342$       &$0.411$       &$0.376$   \\
            Vis-MVSNet \cite{zhang2020visibility}                 &$0.369$       &$0.361$       &$0.365$   \\
            AA-RMVSNet \cite{wei2021aa}                           &$0.376$       &$0.339$       &$0.357$   \\
            CasMVSNet \cite{gu2020cascade}                        &$0.325$       &$0.385$       &$0.355$   \\
            EPP-MVSNet \cite{ma2021epp}                           &$0.413$       &\B{0.296}       &$0.355$   \\
            CVP-MVSNet \cite{yang2020cost}                        &\B{0.296}       &$0.406$       &$0.351$   \\
            UCSNet \cite{cheng2020deep}                           &$0.338$       &$0.349$       &$0.344$   \\
            IB-MVS \cite{sormann2021ib}                             &$0.334$    &$0.309$        &\B{0.321} \\
            \hline
Ours & $0.359$            & $0.305$            & $0.332$             \\ \hline
\end{tabular}
\end{table}

\begin{table*}[]

\caption{Results on \tnt{}}
\label{tab:tnt}
\centering
\resizebox{\textwidth}{!}{
  
\begin{tabular}{llllllllll|lllllll}
\hline
                                                            & \multicolumn{9}{c|}{intermediate}                                                       & \multicolumn{7}{c}{advanced}                                        \\ \cline{2-17} 
Method                                                      & mean    & Fam.    & Franc.  & Horse   & Light.  & M60     & Pan.    & Play.   & Train   & mean    & Audi.   & Ballr.  & Courtr. & Museum  & Palace  & Temple  \\ \hline
COLMAP \cite{schonberger2016pixelwise}     & $42.14$ & $50.41$ & $22.25$ & $25.63$ & $56.43$ & $44.83$ & $46.97$ & $48.53$ & $42.04$ & $27.24$ & $16.02$ & $25.23$ & $34.7$  & $41.51$ & $18.05$ & $27.94$ \\
MVSNet \cite{yao2018mvsnet}                & $43.48$ & $55.99$ & $28.55$ & $25.07$ & $50.79$ & $53.96$ & $50.86$ & $47.90$ & $34.69$ & -       & -       & -       & -       & -       & -       & -       \\
Point-MVSNet \cite{chen2019point}          & $48.27$ & $61.79$ & $41.15$ & $34.20$ & $50.79$ & $51.97$ & $50.85$ & $52.38$ & $43.06$ & -       & -       & -       & -       & -       & -       & -       \\
CVP-MVSNet \cite{yang2020cost}             & $54.03$ & $76.50$ & $47.74$ & $36.34$ & $55.12$ & $57.28$ & $54.28$ & $57.43$ & $47.54$ & -       & -       & -       & -       & -       & -       & -       \\
UCSNet \cite{cheng2020deep}                & $54.83$ & $76.09$ & $53.16$ & $43.03$ & $54.00$ & $55.60$ & $51.49$ & $57.38$ & $47.89$ & -       & -       & -       & -       & -       & -       & -       \\
Altizure-SFM, PCF-MVS \cite{kuhn2019plane} & $55.88$ & $70.99$ & $49.60$  & $40.34$ & $63.44$ & $57.79$ & $58.91$ & $56.59$ & $49.40$  & $35.69$ & $28.33$ & $38.64$ & $35.95$ & $48.36$ & $26.17$ & $36.69$ \\
IB-MVS \cite{sormann2021ib} &           $56.02$ & $75.76$	&$57.65$	&$41.61$	&$55.90$	&$58.09$	&$51.89$	&$59.48$	&$47.73$ &$31.96$	&$22.41$	&$37.00$	&$31.60$	&$41.01$	&$28.20$	&$31.54$ \\
CasMVSNet \cite{gu2020cascade}             & $56.84$ & $76.37$ & $58.45$ & $46.26$ & $55.81$ & $56.11$ & $54.06$ & $58.18$ & $49.51$ & $31.12$ & $19.81$ & $38.46$ & $29.10$  & $43.87$ & $27.36$ & $28.11$ \\
ACMM \cite{xu2019multi}                    & $57.27$ & $69.24$ & $51.45$ & $46.97$ & $63.20$ & $55.07$ & $57.64$ & $60.08$ & $54.48$ & $34.02$ & $23.41$ & $32.91$ & $41.17$ & $48.13$ & $23.87$ & $34.60$  \\
ACMP \cite{xu2020planar}                   & $58.41$ & $70.30$  & $54.06$ & \B{54.11} & $61.65$ & $54.16$ & $57.60$  & $58.12$ & $57.25$ & $37.44$ & \B{30.12} & $34.68$ & \B{44.58} & $50.64$ & $27.20$  & $37.43$ \\
Altizure-HKUST-2019 \cite{Altizure}        & $59.03$ & $77.19$ & $61.52$ & $42.09$ & $63.50$  & $59.36$ & $58.20$  & $57.05$ & $53.3$  & $37.34$ & $24.04$ & $44.52$ & $36.64$ & $49.51$ & $30.23$ & $39.09$ \\
DeepC-MVS \cite{kuhn2020deepc} & $59.79$&	$71.91$&	$54.08$&	$42.29$&	$66.54$&	$55.77$&	\B{67.47}&	$60.47$&	\B{59.83} & $34.54$& $26.30$&	$34.66$&	$43.50$&	$45.66$&	$23.09$&	$34.00$ \\
Vis-MVSNet \cite{zhang2020visibility}      & $60.03$ & $77.40$ & $60.23$ & $47.07$ & $63.44$ & $62.21$ & $57.28$ & $60.54$ & $52.07$ & $33.78$ & $20.79$ & $38.77$ & $32.45$ & $44.20$  & $28.73$ & $37.70$  \\
AttMVS \cite{luo2020attention} & $60.05$ &	$73.90$&	$62.58$&	$44.08$&	$64.88$&	$56.08$&	$59.39$&	$63.42$&	$56.06$& $31.93$&	$15.96$&	$27.71$&	$37.99$&	\B{52.01} &	$29.07$&	$28.84$ \\
D2HC-MVSNet \cite{yan2020dense}            & $60.13$ & $77.36$ & $57.74$ & $45.74$ & $63.39$ & $63.30$ & $57.82$ & $60.71$ & $54.99$ & -       & -       & -       & -       & -       & -       & -       \\
AA-RMVSNet \cite{wei2021aa}                & $61.51$ & $77.77$ & $59.53$ & $51.53$ & $64.02$ & \B{64.05} & $59.47$ & $60.85$ & $54.90$  & -       & -       & -       & -       & -       & -       & -       \\
EPP-MVSNet \cite{ma2021epp}                & $61.68$ & $77.86$ & $60.54$ & $52.96$ & $62.33$ & $61.69$ & $60.34$ & $62.44$ & $55.30$ & $35.72$ & $21.28$ & $39.74$ & $35.34$ & $49.21$ & $30.00$    & $38.75$ \\ \hline
Ours                                                        & \B{64.82}     & \B{81.16}     & \B{64.21}     & $50.43$     & \B{70.73}     & $63.85$     & $63.99$     & \B{65.90}     & $58.25$     & \B{40.19}     & $25.95$     & \B{45.75}     & $39.65$     & $51.75$     & \B{35.08}     & \B{42.97}     \\ \hline

\end{tabular}
}
\end{table*}

\newcommand{\mww} {5cm } 
\begin{figure*}[h]
 \centering
  \begin{subfigure}{.32\textwidth}
  \centering
    \includegraphics[width=\mww]{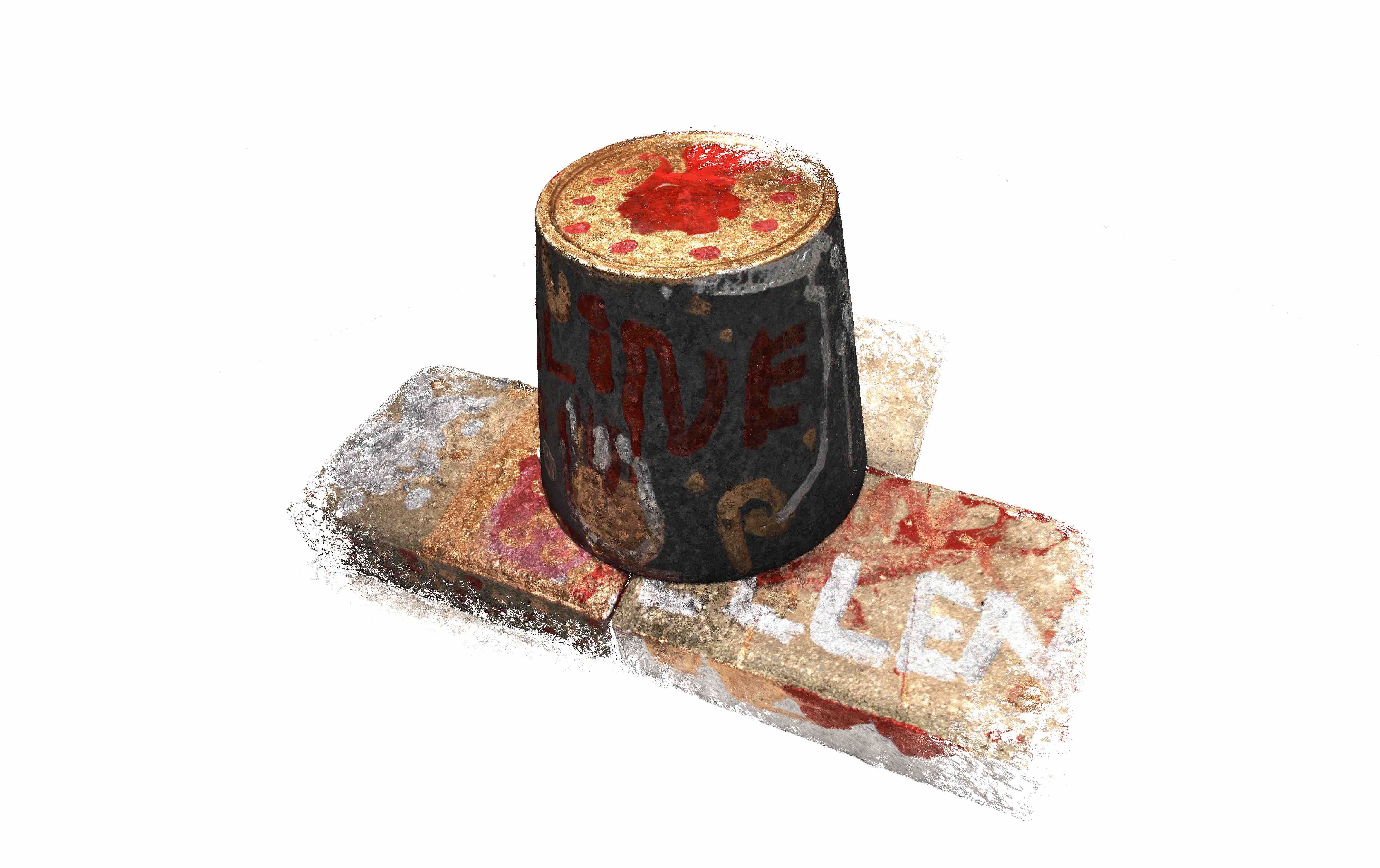}
    \caption{scan1}
  \end{subfigure}
  \centering
    \begin{subfigure}{.32\textwidth}
    \centering
    \includegraphics[width=\mww]{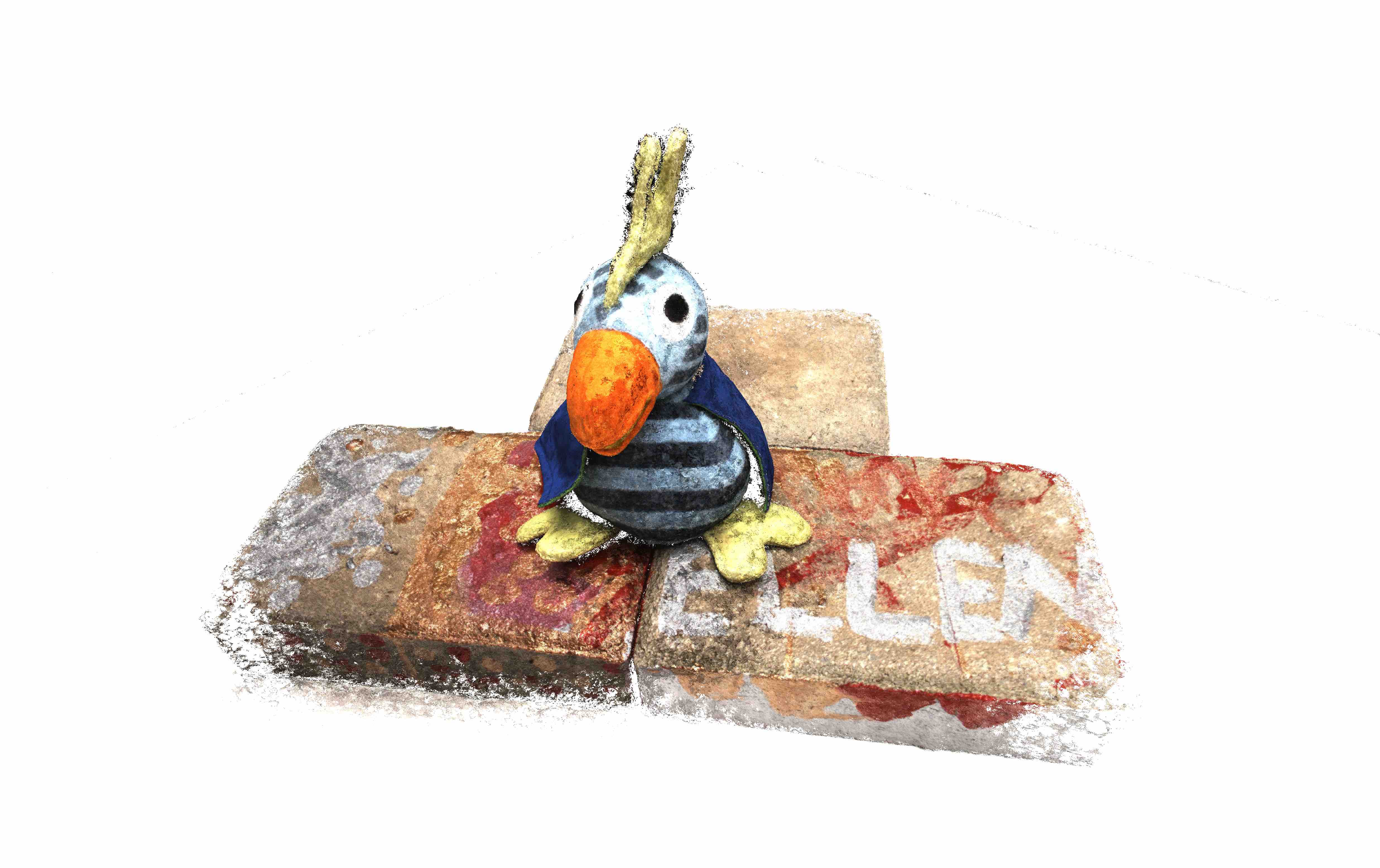}
    \caption{scan4}
  \end{subfigure}
    \begin{subfigure}{.32\textwidth}
  \centering
    \includegraphics[width=\mww]{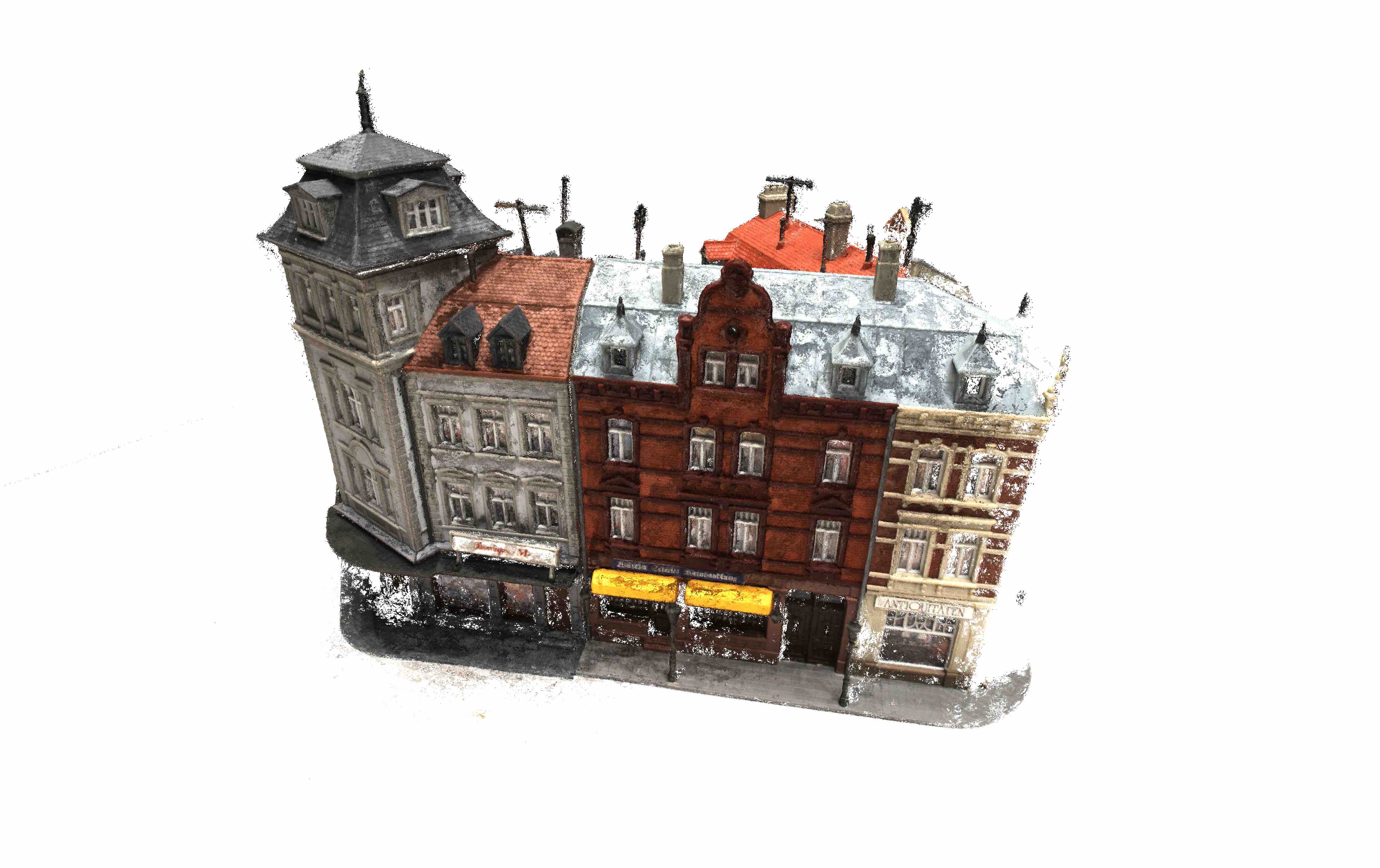}
    \caption{scan9}
  \end{subfigure}
  
  \centering
    \begin{subfigure}{.32\textwidth}
    \centering
    \includegraphics[width=\mww]{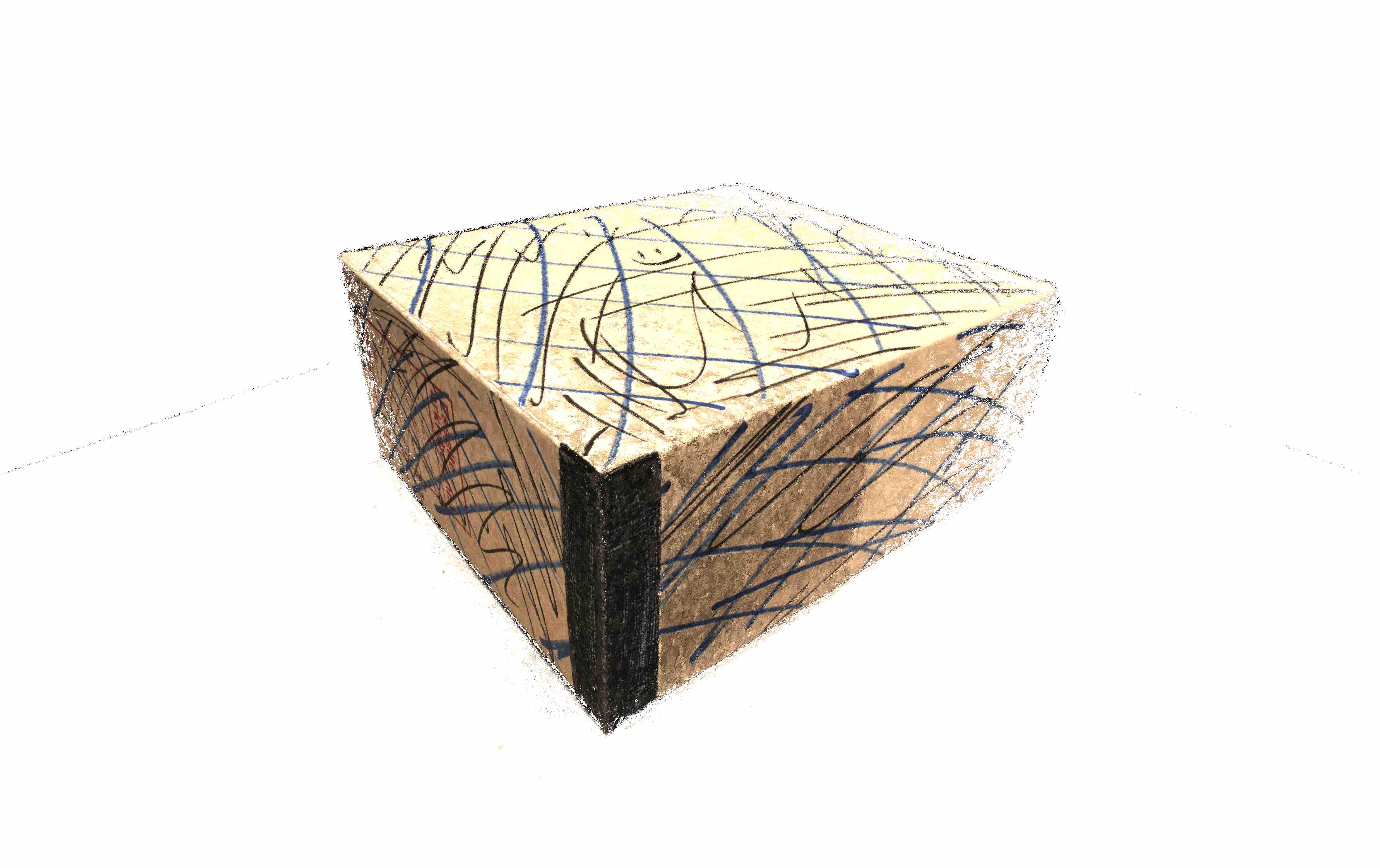}
    \caption{scan10}
  \end{subfigure}
      \begin{subfigure}{.32\textwidth}
    \centering
    \includegraphics[width=\mww]{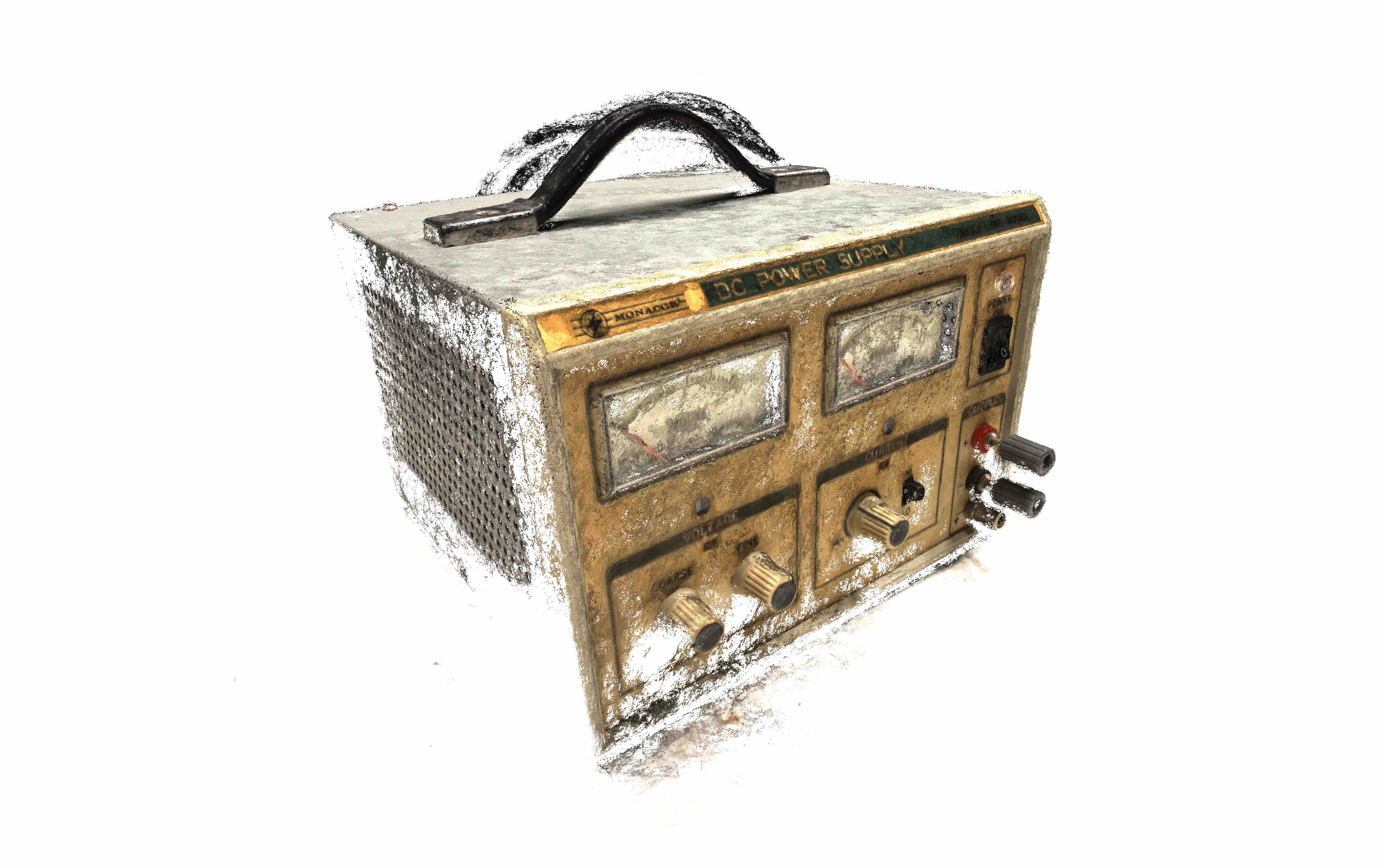}
    \caption{scan11}
  \end{subfigure}
      \begin{subfigure}{.32\textwidth}
    \centering
    \includegraphics[width=\mww]{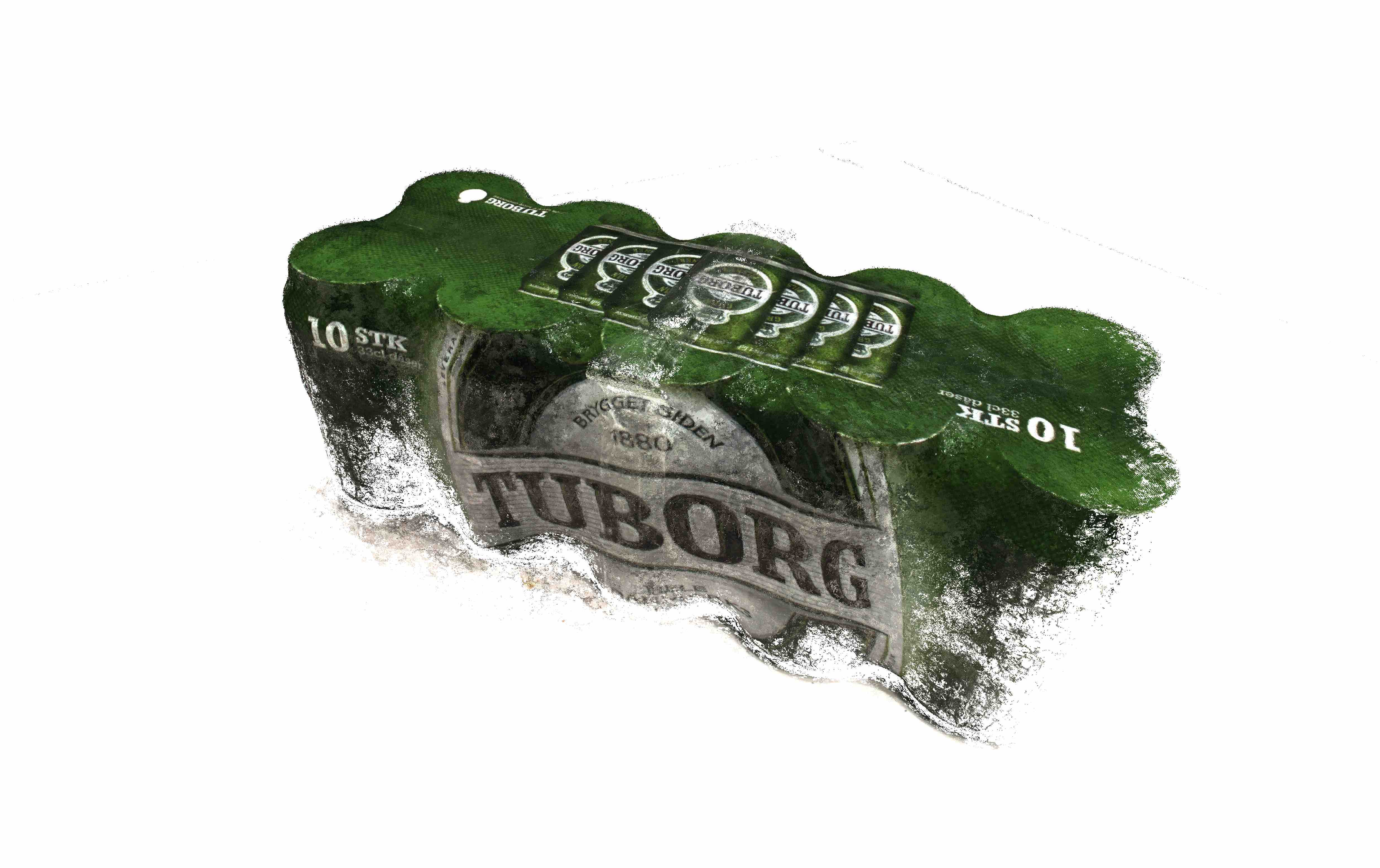}
    \caption{scan12}
  \end{subfigure}
  
  \caption{Visualization of results on \dtu{} (test set).}
  \label{fig:dtu}

 \end{figure*}
 
\begin{figure*}[h]
\newcommand{\mw} {4cm } 
  \begin{subfigure}{0.32\textwidth}
    \centering
    \includegraphics[width=\mw]{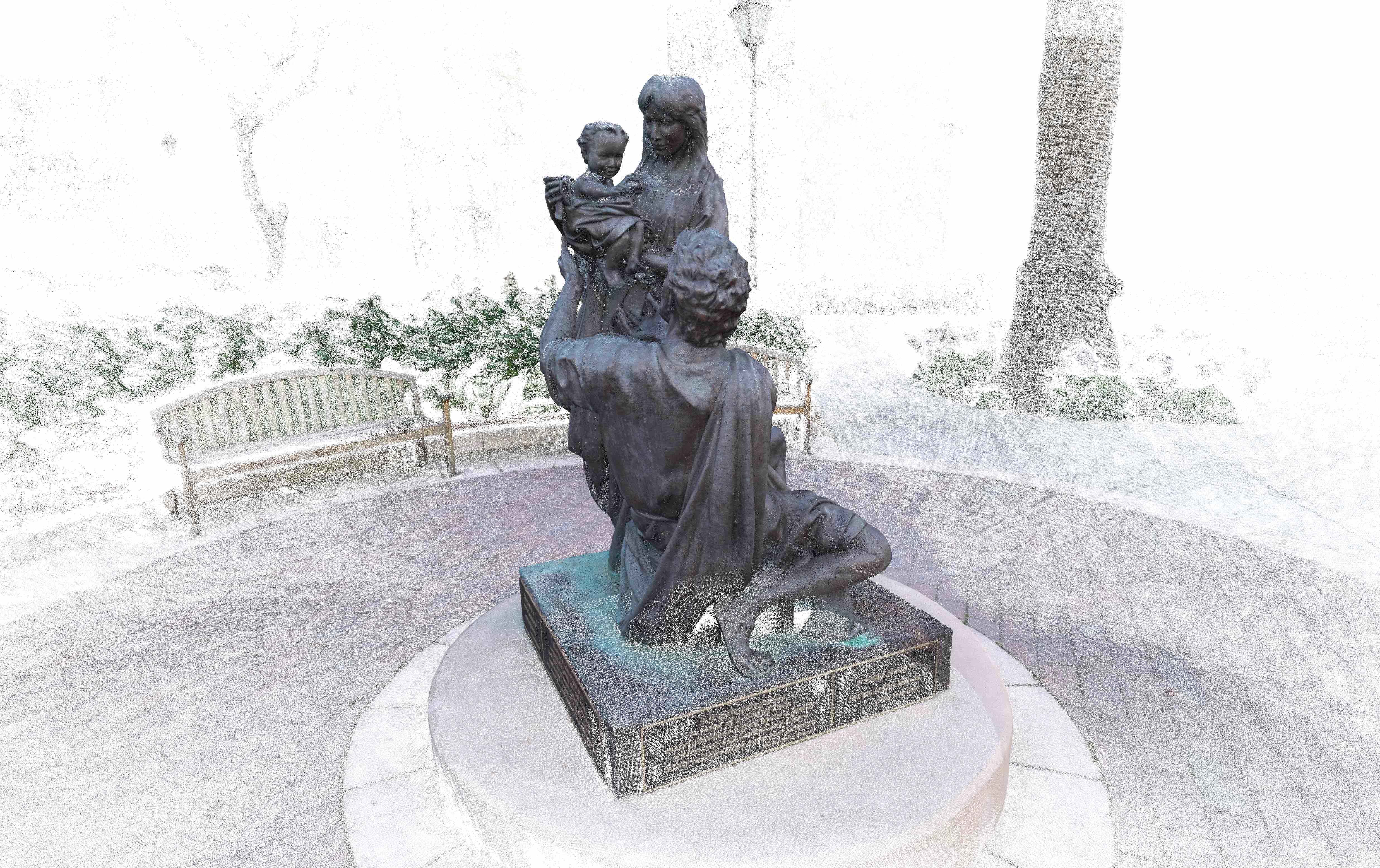}
    \caption{Family (Intermediate)}
  \end{subfigure}
    \begin{subfigure}{0.32\textwidth}
    \centering
    \includegraphics[width=\mw]{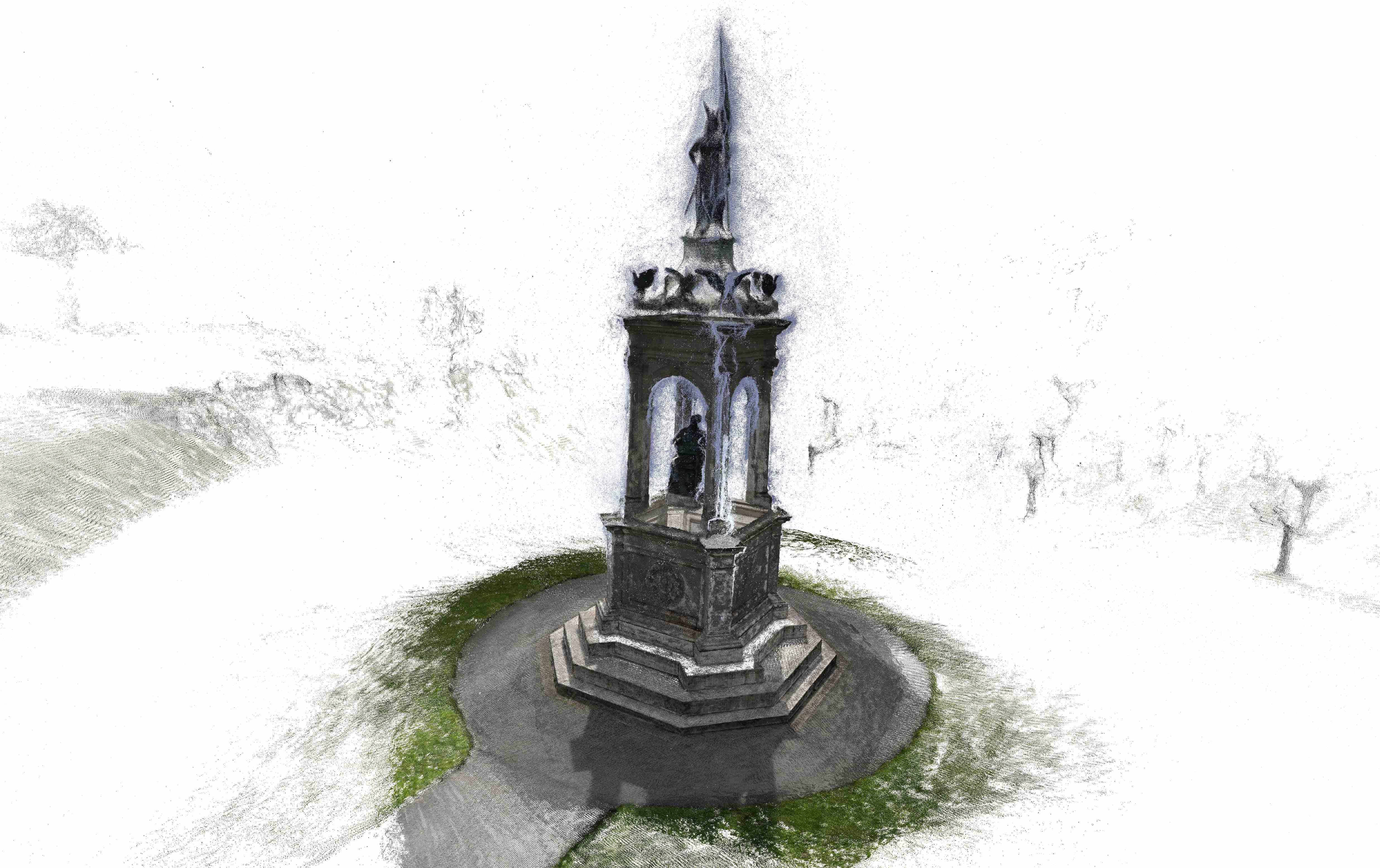}
    \caption{Francis (Intermediate)}
  \end{subfigure}
    \begin{subfigure}{0.32\textwidth}
          \centering
    \includegraphics[width=\mw]{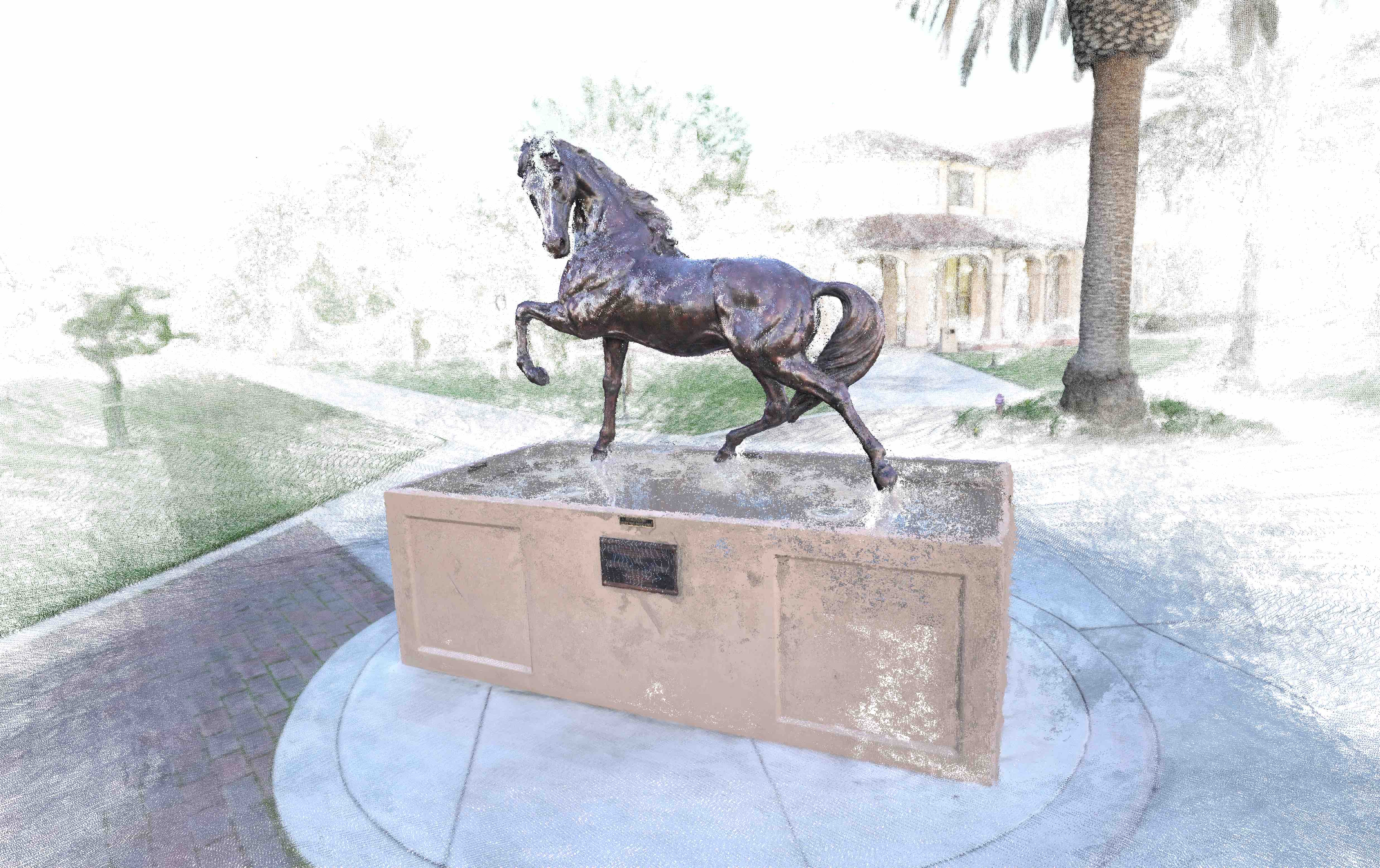}
    \caption{Horse (Intermediate)}
  \end{subfigure}

    \begin{subfigure}{0.32\textwidth}
    \centering
    \includegraphics[width=\mw]{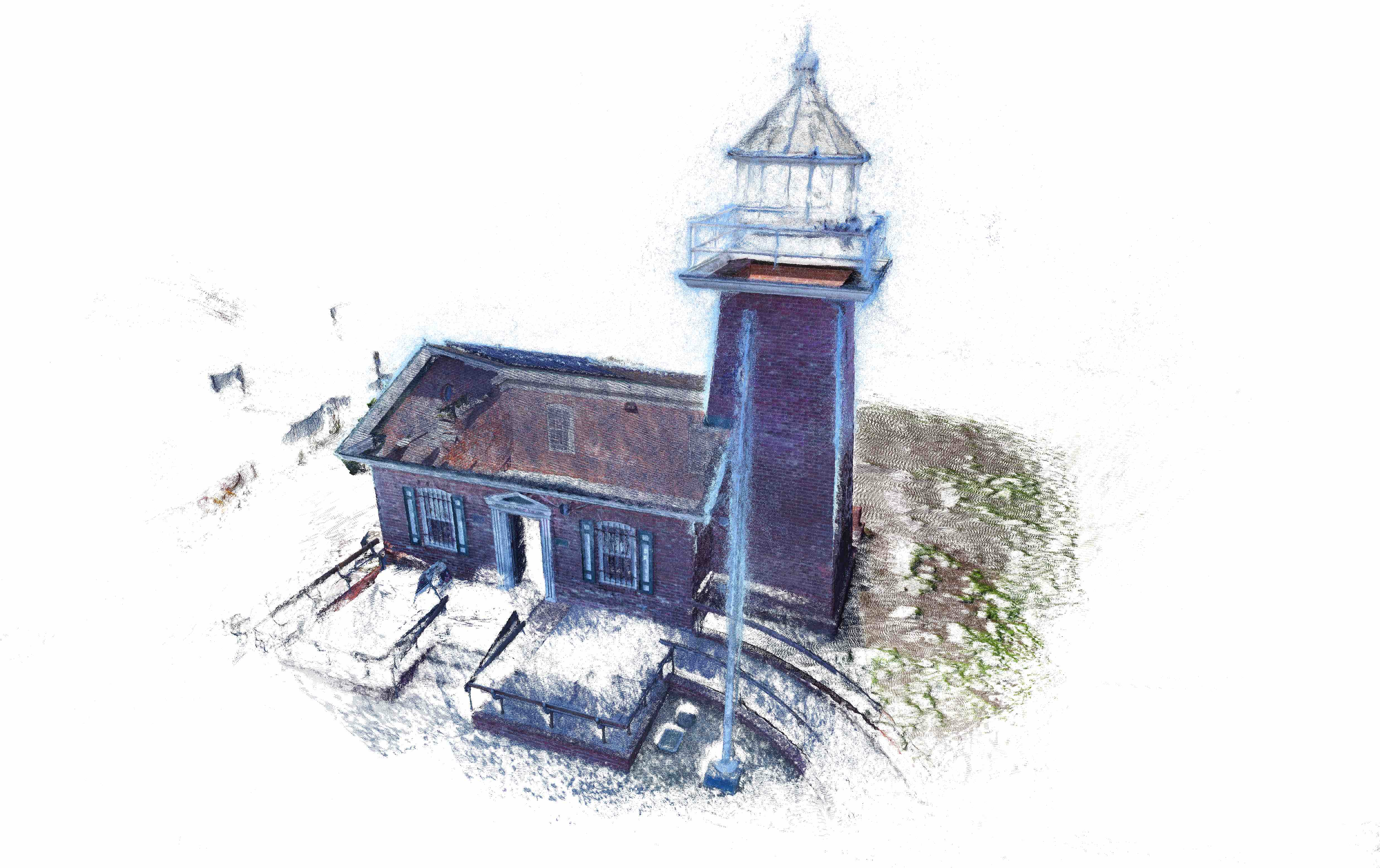}
    \caption{Lighthouse (Intermediate)}
  \end{subfigure}
    \begin{subfigure}{0.32\textwidth}
    \centering
    \includegraphics[width=\mw]{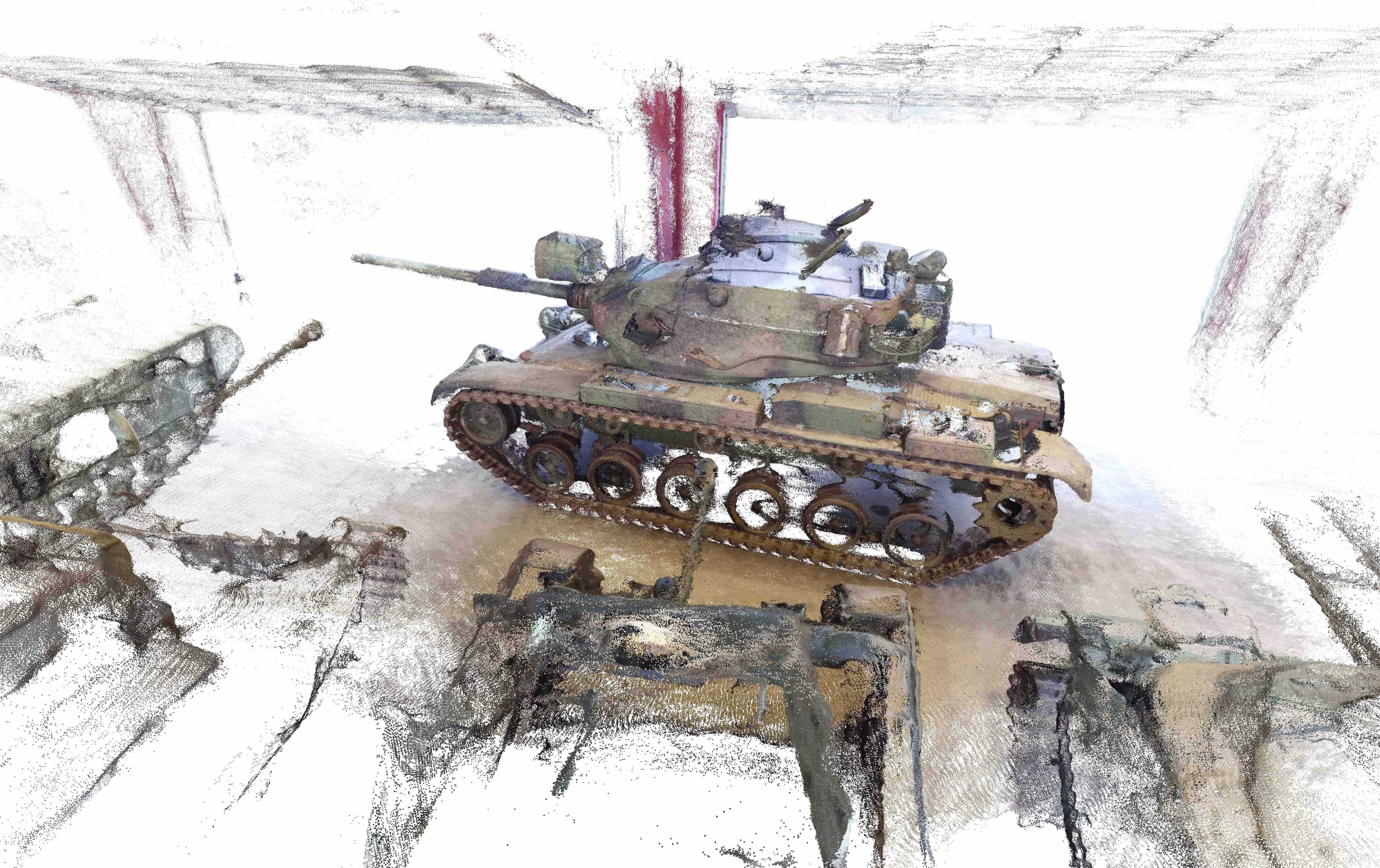}
    \caption{M60 (Intermediate)}
  \end{subfigure}
         \begin{subfigure}{0.32\textwidth}
    \centering
    \includegraphics[width=\mw]{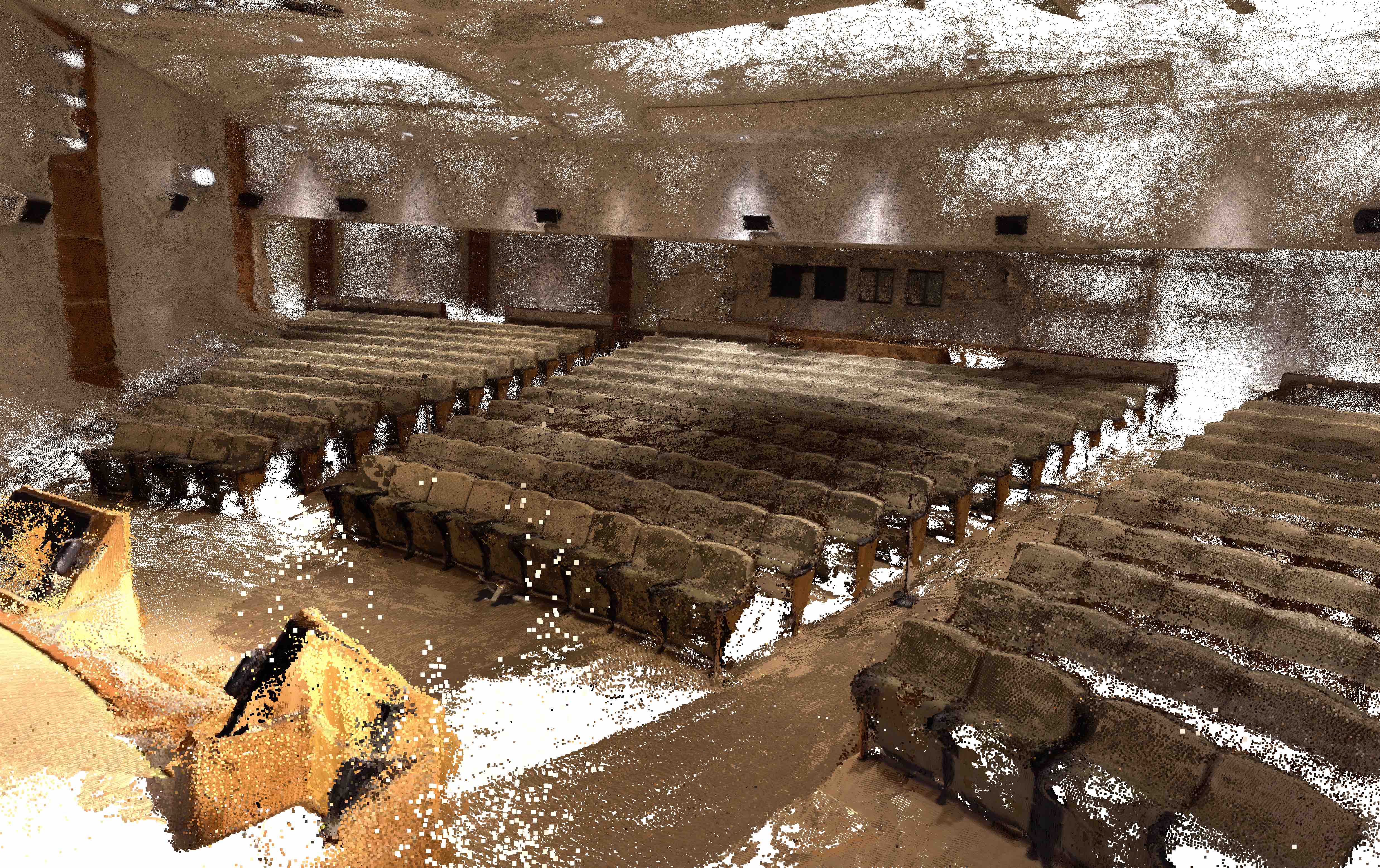}
    \caption{Auditorium (Advanced)}
  \end{subfigure}

    \begin{subfigure}{0.32\textwidth}
    \centering
    \includegraphics[width=\mw]{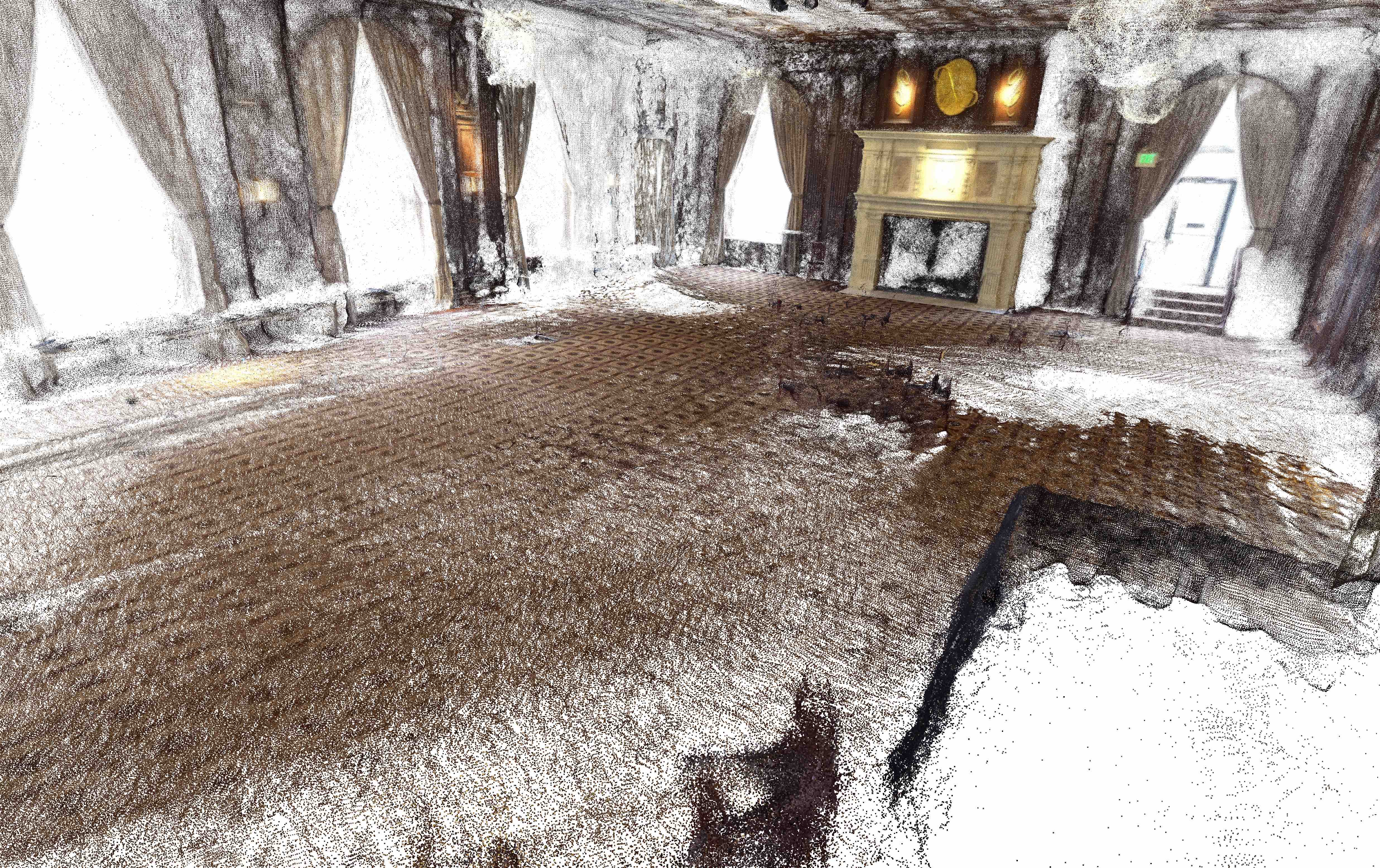}
    \caption{Ballroom (Advanced)}
  \end{subfigure}
    \begin{subfigure}{0.32\textwidth}
          \centering
    \includegraphics[width=\mw]{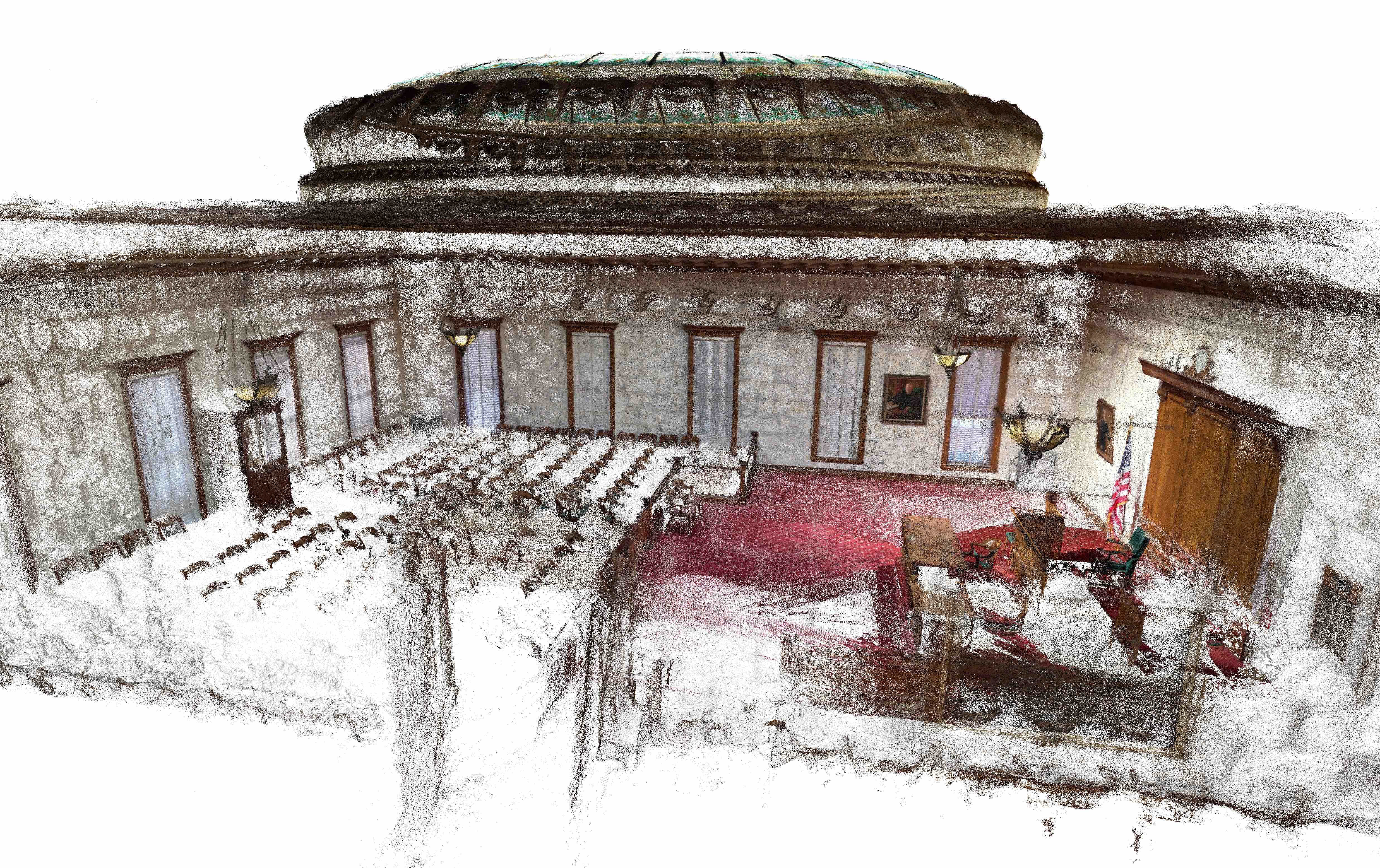}
    \caption{Courtroom (Advanced)}
  \end{subfigure}
      \begin{subfigure}{0.32\textwidth}
          \centering
    \includegraphics[width=\mw]{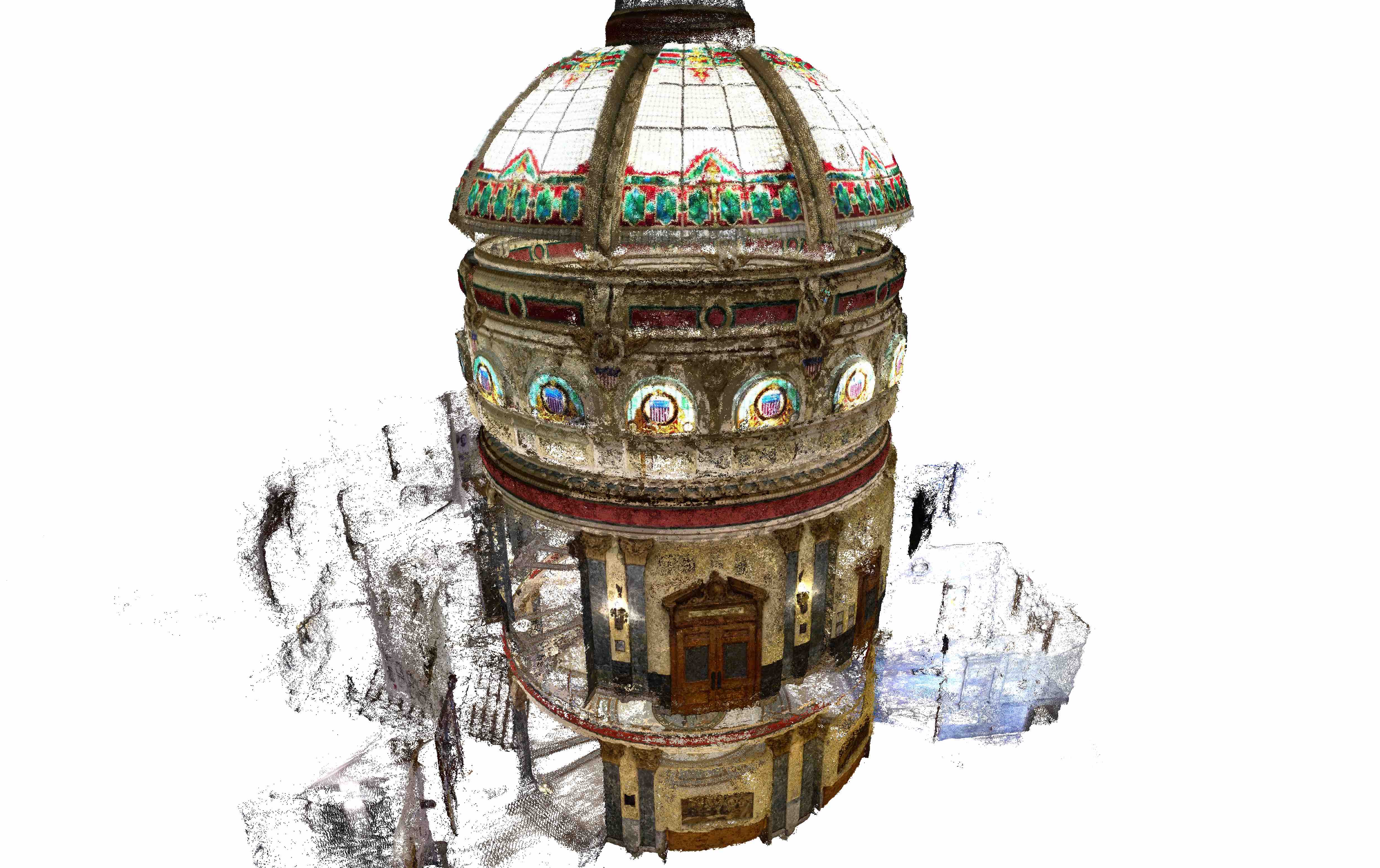}
    \caption{Museum (Advanced)}
  \end{subfigure}

  \caption{Visualization of results on \tnt{}.}
  \label{fig:tnt}
\end{figure*}

\subsection{Adaptive Point Cloud Stitching}

As a last step, the depth maps from the reference views are stitched together to form a single point cloud. We use an adaptive thresholding approach based on Dynamic Consistency Checking (DCC) proposed in D2HC-RMVSNet~\cite{yan2020dense}. DCC hard-codes two thresholds $t_1$ and $t_2$ for reprojection errors, however, we use the thresholds $k t_1$ and $k t_2$ where $k$ is different for each scene to ensure a fixed percentage, $p\%$ of all pixels pass through consistency test. And $p$ is optimized through the validation set.

\subsection{Supervision}
We supervise our network with a loss consisting of two parts. The first part measures the L1 error of the predicted disparity against the ground truth at each iteration, with exponentially increasing weights for later iterations. This part enables faster training of all disparity ranges regardless of outliers at the beginning. The second part of the loss is similar to the first part except that (1) it measures the error of depth (i.e.\@ inverted disparity) so as to be more aligned with point cloud evaluation,  and that (2) the error is capped at a constant $\kappa$ so as to prevent outliers from dominating the loss.

Given the predicted disparity in each iteration be $\disp _t, t=1,..., \Kone+\Ktwo$ and ground truth disparity $\disp _{\rm gt}$, the combined loss is defined as follows:

\begin{align}
\mathcal{L}_1 &=\sum_{t=1}^{\Kone+\Ktwo} \gamma^{\Kone+\Ktwo-t}\left\|\disp _{\rm gt}-\disp _{t}\right\|_{1} \\
\mathcal{L}_2 &=\sum_{t=1}^{\Kone+\Ktwo} \gamma^{\Kone+\Ktwo-t} \min( \left\| \disp_{\rm gt} ^ {-1}- \disp _{t} ^ {-1}\right\|_{1}, {\rm \kappa}) \\
\mathcal{L}   &= (1 - w) \cdot \mathcal{L}_1 + w \cdot \mathcal{L}_2 \cdot \lambda
\end{align}
where $\gamma$ controls the weights across iterations and $\lambda$ makes the two parts have roughly the same range. The parameter $w$ balances the two parts and changes from 0 to 1 linearly as training progresses to focus more on the depth error, e.g.\@ for a total number of 16 training epochs, $w$ would be 0.5 when 8 epochs are finished.

\section{Experiments}

\subsection {Implementation Details}

\label{sec:imp}
We evaluate our models on two datasets, DTU and \tnt{}. On DTU, we train on its training split of DTU and evaluate on its test split, which was suggested by Yao et al.~\cite{yao2018mvsnet} and followed by most authors.
On Tanks-and-Temples,  we train on the \bld{} dataset~\cite{yao2020blendedmvs}, following the practice of prior work ~\cite{yao2018mvsnet,yan2020dense,ma2021epp}. For all datasets, during training we use the native image resolutions after some random cropping and scaling as input to the network
and other details on the  hyperparameters are given in Table \ref{tab:imp}.

To pair neighbor views with reference views, we use the same method as MVSNet ~\cite{yao2018mvsnet}. In \bld{}, which is used for training only, the scenes have large variations in the range of depth values, we scale each reference view, along with its neighbor views, so that its ground-truth depth has a median value $600$ mm. When we evaluate on \tnt{}, due to lack of ground-truth and noisy background, we scale each reference view, along with its neighbor views, so that its minimum depth of a set of reliable feature points (computed by COLMAP ~\cite{schonberger2016pixelwise} as in MVSNet~\cite{yao2018mvsnet}) is $400$ mm. To stitch the predicted depth maps from multiple reference views, we simply scale back each depth map to its original scale.

\subsection{Main Results}

 \paragraph{DTU}
 The results on the DTU benchmark are presented in Table \ref{tab:dtu}. Our method achieves the second best overall score, which is an average completeness and accuracy~\cite{aanaes2016large}. Visualizations of sample reconstructions on DTU are shown in Fig.~\ref{fig:dtu}.

\paragraph{Tanks-and-Temples}
On the \tnt{} dataset, we achieve state of the art performance, as shown in Table \ref{tab:tnt}. Notably, the model is trained on the \bld{} dataset  without finetuning on \tnt{} except for some test-time hyperparameter selection using the validation set,  as described in Table \ref{tab:imp}. This indicates a good generalization ability of our approach.
A visualization of some results is shown in Fig. \ref{fig:tnt}, from which we can see that many reconstructed scenes look reasonably accurate, detailed, and complete, but there is still substantial room for improvement, especially on low-texture planar regions.

\subsection {Ablations}

\label{sec:abl}

We show our ablation experiments on \tnt{} official training set (used as validation set) in a restricted setting where we only train the model on \bld{} for 2 epochs but keep everything else the same as in Table \ref{tab:imp}.

\begin{figure*}
  \centering
    \includegraphics[width=0.9\linewidth]{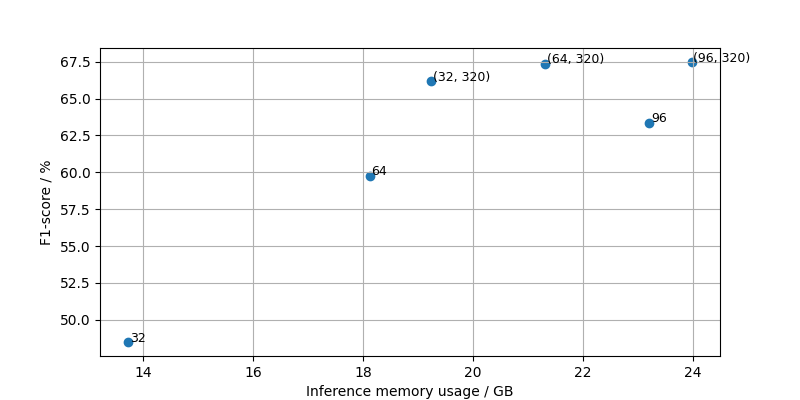}
  \caption{Memory usage of cascaded v.s. non-cascaded model. The label for cascaded models means coarse and fine disparity increments (larger number means smaller increments), and the label for non-cascaded models means the single disparity increment.}
  \label{fig:cas}
\end{figure*}

\paragraph{Cost Volume Cascading}

We study the effect of cost volume cascading on memory consumption.
In Fig.~\ref{fig:cas}, we plot the GPU memory usage versus $F_1$-score on \tnt{} validation set for (1) a series of cascaded model (with different disparity increments in the first stage), (2) its non-cascaded counterpart, which matches the first-stage disparity resolution used in the cascaded model and has equal total GRU iterations. 
We train all models as described in Sec.~\ref{sec:imp} and finally chose the cascaded model (64, 320) for long-time training and benchmarking.
It uses 44 disparity values with an increment of $d_{\rm max}/320$ in the second stage, and uses 64 values with a coarser increment $d_{\rm max}/64$ in the first stage to cover the entire disparity range from $0$ to $d_{\rm max}$.
For the non-cascaded model, because it needs to fill the entire disparity range from $0$ to $d_{\rm max}$, it needs significantly more memory as the disparity resolution increases. 
We see from Fig.~\ref{fig:cas} that cascading produces significant savings of memory.
Note the reported memory is the peak memory reported by the command "nvidia-smi".

 \begin{table}

       \centering
  \caption{Ablation on supervision}
    \label{tab:sup}

\begin{tabular}{l|l}
\hline
Method & $F_1$-score  \\ \hline
(1) Truncated $L_1$   depth loss & N/A   \\ \hline
(2) $L_1$  disparity loss         & $66.79$ \\ \hline
(3) Average of (1) and (2)    & $67.32$ \\ \hline
(4) Proposed dynamic loss     & \B{67.36} \\ \hline
\end{tabular}

   \caption{Ablation of neighbor view number}
  
  \label{tab:nf_tnt}
  \centering

\begin{tabular}{cl|lll}
\hline
\multicolumn{2}{c|}{\multirow{2}{*}{Mean $F_1$-score (\%)}}                                                                          & \multicolumn{3}{c}{\begin{tabular}[c]{@{}c@{}}\# Neighbor views\\ in 2$\times$ native resolution\end{tabular}} \\ \cline{3-5} 
\multicolumn{2}{c|}{}                                                                                                             & $5$                                 & $15$                                & $25$                               \\ \hline
\multicolumn{1}{c|}{\multirow{3}{*}{\begin{tabular}[c]{@{}c@{}}\# Neighbor views\\ in\\  native resolution\end{tabular}}} & $5$  & $62.62$                             & $66.42$                             & $67.27$                            \\
\multicolumn{1}{c|}{}                                                                                                        & $15$ & $62.73$                             & $66.48$                             & \B{67.36}                            \\
\multicolumn{1}{c|}{}                                                                                                        & $25$ & $62.66$                             & $66.37$                             & $67.27$                            \\ \hline
\end{tabular}

\end{table}

\paragraph {Dynamic Supervision}
In Table \ref{tab:sup}, we show our model trained with different loss supervision. Among them, the truncated $L_1$ depth loss does not help the model to start up; and $L_1$ disparity loss has inferior performance; while the proposed dynamic loss is marginally better than the direct average of $L_1$ depth loss and $L_1$ disparity loss.

\paragraph {Number of Neighbor Views}
During inference, our network can use a different number of neighbor views than in training. 
In table \ref{tab:nf_tnt}, we study the effect of changing the number of neighbor views during inference. In particular, we study how this number can be chosen differently for the two resolutions we use to predict depth maps. 
As the results on the validation set show, the best combination is 15 views for native resolution prediction and 25 views for 2 $\times$ native resolution prediction. And these are the numbers we use on the test set.

 \begin{table}[!ht]

  \caption{Ablation of aggregation options}
  
  \label{tab:agg}
  \centering

\begin{tabular}{l|l}
\hline
Aggergation option & Mean $F_1$-score (\%) \\ \hline
max                & $57.77$                  \\ \hline
max + mean         & $65.37$                  \\ \hline
std                & $59.90$                  \\ \hline
std + mean         & $66.85$                  \\ \hline
mean    & \B{67.36}              \\ \hline
\end{tabular}

   \caption{Ablation of adaptive thresholding}
  
  \label{tab:ada_thre}
  \centering
  
      \resizebox{\linewidth}{!}{
\begin{tabular}{l|ccccc}
\hline
Controlled   percentage $p \% $ & $15\%$  & $20\%$  & $25\%$  & $30\%$  & $35\%$  \\ \hline
Mean $F_1$-score (\%)                        & $66.83$ & $67.31$ & \B{67.36} & $67.11$ & $66.60$ \\ \hline

\end{tabular}
}

      \resizebox{\linewidth}{!}{

\begin{tabular}{l|ccccc}
\hline
Fixed threshold  $k$       & $1$     & $1.5$   & $2$     & $2.5$   & $3$     \\ \hline
Mean $F_1$-score (\%)                        &   $65.33$    & $66.10$  & $66.33$ & $66.32$ & $66.13$ \\ \hline
\end{tabular}
}

\end{table}

\paragraph {Aggregation of Cost Volumes}
Here in Table \ref{tab:agg} we study the effect of aggregation options different from our simple averaging including both one-channel and two-channel ones. It shows that taking the mean is the best.

\paragraph {Adaptive Thresholding}
To strike a balance between accuracy and completeness scores, we use adaptive thresholding method and search for the best parameter $p$. The results are in Table \ref{tab:ada_thre} in comparison with results from fixed thresholds. We see that our adaptive thresholding approach is significantly better than fixed thresholding.

 \begin{table*}[!ht]

  \caption{Ablation of multiresolution fusion}
  
  \label{tab:mr}
  \centering
        \resizebox{0.6\textwidth}{!}{
\begin{tabular}{l|ccccc}
\hline
\begin{tabular}[c]{@{}l@{}}Multi-resolution \\ with control threshold $t$\end{tabular} & \begin{tabular}[c]{@{}c@{}}0\\ =native input\end{tabular} & 0.01 & 0.02 & 0.04 & \begin{tabular}[c]{@{}c@{}} $\infty$ \\ =2$\times$native input\end{tabular} \\ \hline
Mean $F_1$-score (\%)                                                              & $64.38$                                                         & $68.47$    & \B{68.49}    & $68.39$    & $68.08$                                                             \\ \hline
\end{tabular}
}

\resizebox{0.6\textwidth}{!}{

\begin{tabular}{l|ccccc}
\hline
Weighted average with $w$ & \multicolumn{1}{c}{\begin{tabular}[c]{@{}c@{}}0\\ =native input\end{tabular}} & 0.25 & 0.5 & 0.75 & \multicolumn{1}{c}{\begin{tabular}[c]{@{}c@{}}1\\ =2$\times$native input\end{tabular}} \\ \hline
Mean F1-score (\%) & $64.38$                                                                                & $65.30$   & $66.55$    &     $67.51$ &     $68.08$                                                                       \\ \hline
\end{tabular}
}

\end{table*}

\paragraph {Multiresolution Fusion of Depth Maps}
\label {sec:ab_mr}
An important part of \methodname{} is the multiresolution fusion of depth maps. Different from previous components, its effect is most obvious on our final model trained for 16 epochs. We report the following results on the validation sets of \tnt{}: (1) Different control parameter $t$, and (2) simple weighted average of native input results and 2 $\times$ native input results with weight $w$ . We see from Table \ref{tab:mr} that our novel fusion approach is significantly better than all the other approaches.

 \begin{table*}[!ht]

  \caption{Comparison of running time and memory cost}
  
  \label{tab:cost}
  \centering
    \resizebox{0.6\textwidth}{!}{
\begin{tabular}{l|c|c|c|cc}
\hline
Method        & \begin{tabular}[c]{@{}l@{}}\# Neighbor\\ views\end{tabular} & \begin{tabular}[c]{@{}l@{}}Input\\ Resolution\end{tabular}              & \begin{tabular}[c]{@{}l@{}}Output\\ Resolution\end{tabular} & \begin{tabular}[c]{@{}l@{}}Times per\\ view (ms)\end{tabular} & \begin{tabular}[c]{@{}l@{}}Mem.\\ (GB)\end{tabular} \\ \hline
CasMVSNet     & \multirow{6}{*}{4}                                          & \multirow{5}{*}{\begin{tabular}[c]{@{}l@{}}(1056,\\ 1920)\end{tabular}} & (1056, 1920)                                                & 792.2                                                         & 9.5                                                 \\ \cline{1-1} \cline{4-6} 
Vis-MVSNet    &                                                             &                                                                         & (528, 960)                                                  & 864.2                                                         & 4.5                                                 \\ \cline{1-1} \cline{4-6} 
PatchmatchNet &                                                             &                                                                         & (1056, 1920)                                                & 317.7                                                         & 3.2                                                 \\ \cline{1-1} \cline{4-6} 
EPP-MVSNet    &                                                             &                                                                         & (528, 960)                                                  & 522.2                                                         & 8.2                                                 \\ \cline{1-1} \cline{4-6} 
Ours          &                                                             &                                                                         & (264, 480)                                                  & 664.4                                                        & 3.0                                                \\ \cline{1-1} \cline{3-6} 
Ours          &                                                             & \multirow{2}{*}{\begin{tabular}[c]{@{}l@{}}(2112,\\ 3840)\end{tabular}} & \multirow{2}{*}{(528, 960)}                                 & 1754.5                                                        & 7.0                                                \\ \cline{1-2} \cline{5-6} 
Ours          & 25                                                          &                                                                         &                                                             & 7611.3                                                        & 22.6                                                \\ \hline
\end{tabular}

}
\end{table*}

\subsection {Memory and Runtime}

The computational cost of \methodname{} is compared with other methods in Table \ref{tab:cost}. When using similar resolution and numbers of views, the time and memory cost of our method is comparable to others.

\section{Conclusion}

\label{sec:con}

We have proposed \methodname{}, a new approach based on the RAFT architecture developed for optical flow. \methodname{} introduces five new changes to RAFT: epipolar cost volumes, cost volume cascading, multiview fusion of cost volumes, dynamic supervision, and multiresolution fusion of depth maps, as well as adaptive thresholding to construct point clouds. Experiments show that our approach achieves competitive performance on \dtu{} and state-of-the-art performance on \tnt{}.

\noindent
\textbf{Acknowledgments: } This work is partially supported by the National Science Foundation under Award IIS-1942981.

{\small
\bibliographystyle{ieee_fullname}
\bibliography{egbib}
}

\onecolumn

\begin{appendices}

\section{Network Architecture}
\label{appA}
 \begin{figure}[h]
  \centering
    \includegraphics[width=\textwidth]{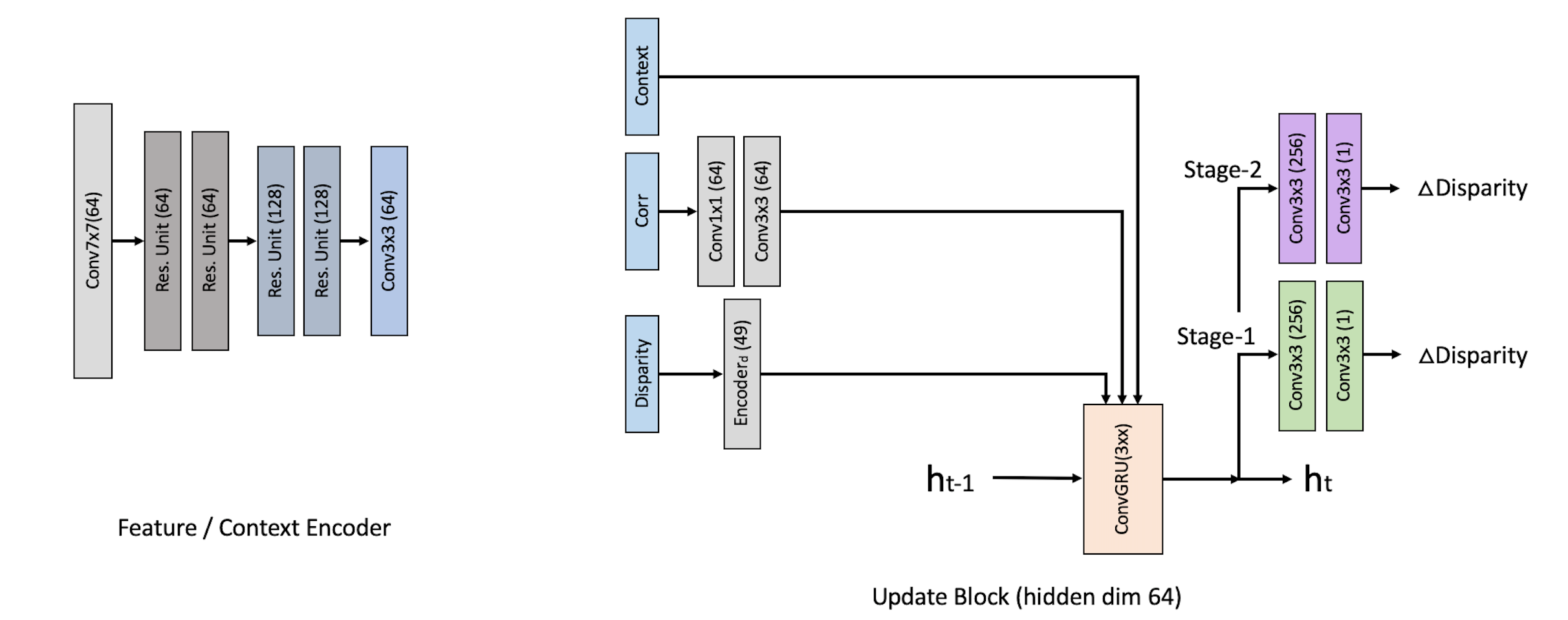}
  \caption{Network architecture details. The context and feature encoders have the same architecture, the only difference is that the feature encoder uses instance normalization while the context encoder uses batch normalization.}
  \label{fig:arch}
\end{figure}

\end{appendices}

\end{document}